\documentclass{article}

\usepackage{amsmath,amsfonts,bm}

\def\1{\bm{1}}

\def\rr{{\textnormal{r}}}

\DeclareMathAlphabet{\mathsfit}{\encodingdefault}{\sfdefault}{m}{sl}
\SetMathAlphabet{\mathsfit}{bold}{\encodingdefault}{\sfdefault}{bx}{n}

\usepackage[utf8]{inputenc} %
\usepackage[T1]{fontenc} %
\usepackage{hyperref}
\usepackage{url}
\usepackage{booktabs}
\usepackage{nicefrac} %
\usepackage{microtype} %

\usepackage{wrapfig} %
\usepackage{graphicx}

\usepackage{amsmath}
\usepackage{amssymb}
\usepackage{algorithm} %
\usepackage{algorithmic} %
\usepackage{enumerate} %
\usepackage{multirow} %
\usepackage{amsmath}
\usepackage{subfigure}
\usepackage{helvet} %
\usepackage{courier} %
\usepackage{color}
\usepackage[accepted]{icml2020}

\newcommand{\Wmat}[0]{\ensuremath{{\bf W}} }
\newcommand{\Xmat}[0]{\ensuremath{{\bf X}} }

\newcommand{\av}[0]{\ensuremath{\boldsymbol{a}} }
\newcommand{\bv}[0]{\ensuremath{\boldsymbol{b}} }

\newcommand{\dv}[0]{\ensuremath{\boldsymbol{d}} }
\newcommand{\ev}[0]{\ensuremath{\boldsymbol{e}} }

\newcommand{\hv}[0]{\ensuremath{\boldsymbol{h}} }

\newcommand{\kv}[0]{\ensuremath{\boldsymbol{k}} }

\newcommand{\rrv}[0]{\ensuremath{\boldsymbol{r}} }
\newcommand{\sv}[0]{\ensuremath{\boldsymbol{s}} }
\newcommand{\xv}[0]{\ensuremath{\boldsymbol{x}} }

\newcommand{\zv}[0]{\ensuremath{\boldsymbol{z}} }
\newcommand{\Av}[0]{\ensuremath{\boldsymbol{A}} }

\newcommand{\Zv}[0]{\ensuremath{\boldsymbol{Z}} }
\newcommand{\Pimat}[0]{\ensuremath{\boldsymbol{\Pi}} }
\newcommand{\Phimat}[0]{\ensuremath{\boldsymbol{\Phi}}}
\newcommand{\Omegamat}[0]{\ensuremath{\boldsymbol{\Omega}}}

\newcommand{\epsilonv}[0]{\ensuremath{\boldsymbol{\epsilon}} }

\newcommand{\etav}[0]{\ensuremath{\boldsymbol{\eta}} }
\newcommand{\thetav}[0]{\ensuremath{\boldsymbol{\theta}} }

\newcommand{\lambdav}[0]{\ensuremath{\boldsymbol{\lambda}} }

\newcommand{\piv}[0]{\ensuremath{\boldsymbol{\pi}} }
\newcommand{\phiv}[0]{\ensuremath{\boldsymbol{\phi}} }
\newcommand{\cdotv}[0]{\ensuremath{\boldsymbol{\cdot}}}
\newcommand{\mc}{\multicolumn}
\newcommand{\mr}{\multirow}
\newcommand{\given }{\,|\,}

\def\rr{\textcolor{red}}

\icmltitlerunning{Recurrent Hierarchical Topic-Guided RNN for Language Generation}
\begin{document}

\twocolumn[
\icmltitle{Recurrent Hierarchical Topic-Guided RNN for Language Generation}
\begin{icmlauthorlist}
\icmlauthor{Dandan Guo}{to}
\icmlauthor{Bo Chen}{to}
\icmlauthor{Ruiying Lu}{to}
\icmlauthor{Mingyuan Zhou}{ed}
\end{icmlauthorlist}

\icmlaffiliation{to}{National Laboratory of Radar Signal Processing, Xidian University, Xi'an, China.}
\icmlaffiliation{ed}{McCombs School of Business, The University of Texas at Austin, Austin, TX 78712, USA}

\icmlcorrespondingauthor{Bo Chen}{bchen@mail.xidian.edu.cn}

\icmlkeywords{Machine Learning, ICML}

\vskip 0.3in
]

\printAffiliationsAndNotice{}

\begin{abstract}
To simultaneously capture syntax and global semantics from a text corpus, we propose a new larger-context recurrent neural network (RNN) based language model, which extracts recurrent hierarchical semantic structure via a dynamic deep topic model to guide natural language generation. Moving beyond a conventional RNN-based language model that ignores long-range word dependencies and sentence order, the proposed model captures not only intra-sentence word dependencies, but also temporal transitions between sentences and inter-sentence topic dependencies. For inference, we develop a hybrid of stochastic-gradient Markov chain Monte Carlo and recurrent autoencoding variational Bayes. Experimental results on a variety of real-world text corpora demonstrate that the proposed model not only outperforms larger-context RNN-based language models, but also learns interpretable recurrent multilayer topics and generates diverse sentences and paragraphs that are syntactically correct and semantically coherent. 
\end{abstract}

\section{Introduction}
Both topic and language models are widely used for text analysis.
Topic models, such as latent Dirichlet allocation (LDA) \citep{blei2003latent,griffiths2004finding,hoffman2013stochastic} and its nonparametric Bayesian generalizations %
\citep{Teh2006Hierarchical,NBP2012}, %
 are well suited for extracting document-level word concurrence patterns into latent topics from a text corpus. %
 Their modeling power has been further enhanced by introducing multilayer deep representation
 \citep{srivastava2013modeling,mnih2014neural,gan2015scalable,GBN,zhao2018dirichlet,Zhang2018WHAI}.
 While %
 having semantically meaningful latent representation, they typically treat each document as a bag of words (BoW), ignoring word order \citep{griffiths2004integrating,wallach2006topic}. To take the word order into consideration, \citet{wang2019convolutional} introduce a customized convolutional operator and probabilistic pooling into a topic model, %
 which successfully captures local dependencies and forms phrase-level topics but has limited ability in modeling sequential dependencies and generating word sequences. 
 
Language models have become key components of various natural language processing tasks, such as text summarization \citep{rush2015a,gehrmann2018bottom}, speech recognition \citep{mikolov2010recurrentNEW,alex2013}, machine translation \citep{sutskever2014sequence,Cho2014Learning}, and image captioning \citep{vinyals2015show,mao2015deep,xu2015show,gan2017semantic,rennie2017self,Fan2020Adaptive}.
The primary purpose of a language model is to capture the distribution of a word sequence, commonly with a recurrent neural network (RNN) \citep{mikolov2011extensions,graves2013generating} {or a Transformer-based model \citep{vaswani2017attention,dai2019transformerxl,devlin2019bert,radford2018improving,radford2019language}. In this paper, utilizing a deep dynamic model for sequentially observed count vectors and introducing a recurrent variational inference network, we focus on improving RNN-based language models that often have much fewer parameters and are easier to perform end-to-end training.}

While RNN-based language models do not ignore word order, they often assume that the sentences of a document are independent of each other. This simplifies the modeling task to independently assigning probabilities to individual sentences, ignoring their order and document context \citep{Tian2016Larger}. Such {language} models may consequently fail to capture the long-range dependencies and
{global semantic meaning} of a document \citep{dieng2017topicrnn,wang2018topic}. While a naive solution to explore richer contextual information is to concatenate all previous sentences into a single ``sentence'' and use it as the input of an RNN-based language model, in practice, the length of that sentence is limited given %
the constraint of memory and computation resource. Even if making the length very long, 
this naive solution rarely works well enough to satisfactorily address 
the long-standing research problem of capturing long-range dependencies, 
motivating a variety of more sophisticated methods to improve existing language models %
\citep{dieng2017topicrnn,lau2017topically,wang2018topic,wang2019topic_guided,dai2019transformerxl}. 
Moreover, %
such a solution often clearly enlarges the model size, increasing the difficulty of optimization and risk of overfitting \citep{dieng2017topicrnn}. %
Finding better ways to model long-range dependencies in language modeling is therefore an open research challenge. To relax the sentence independence assumption in language modeling,
 \citet{Tian2016Larger} propose larger-context language models that model the context of a sentence by representing its preceding sentences as either a single or a sequence of BoW vectors, %
 which are then fed directly into the sentence modeling RNN. %

Since topic models are well suited for capturing long-range dependencies, an alternative approach attracting significant recent interest is leveraging topic models to improve {RNN-based} language models.
 \citet{mikolov2012context} use pre-trained topic model features as an additional input to the RNN hidden states and/or output.
\citet{dieng2017topicrnn} and \citet{ahn2017a} combine the predicted word distributions, given by both a topic model and a language model, under variational autoencoder \citep{kingma2013auto}.
\citet{lau2017topically} introduce an attention based convolutional neural network to extract semantic topics, which are used to extend the RNN cell.
{\citet{wang2018topic} learn the global semantic coherence of a document via a neural topic model and use the learned latent topics to build a mixture-of-experts language model. \citet{wang2019topic_guided} further specify a Gaussian mixture model as the prior of the latent code in variational autoencoder, where each mixture component corresponds to a topic.}

While clearly improving the performance of the end task, these existing topic-guided methods still have clear limitations.
For example, they only utilize shallow topic models with only a single stochastic hidden layer for data generation. %
Note several neural topic models use deep neural networks to construct their variational encoders, but still use shallow generative models {as decoders} \cite{miao2017discovering,srivastava2017autoencoding}.
Another key limitation lies in ignoring the sentence order, as each document is treated as a bag of sentences.
Thus once the topic weight vector learned from the document context is given, the task is often reduced to independently assigning probabilities to individual sentences \citep{lau2017topically,wang2018topic,wang2019topic_guided}.

In this paper, as depicted in Fig. \ref{fig:generative_model}, we propose to use recurrent gamma belief network (rGBN) to guide a stacked RNN for language modeling. We refer to the model as rGBN-RNN, %
which integrates rGBN \citep{guo2018deep}, a deep recurrent topic model, and stacked RNN \citep{graves2013generating,chung2017hierarchical}, a neural language model, into a novel larger-context {RNN-based} language model. It simultaneously learns a deep recurrent topic model, extracting document-level multi-layer word concurrence patterns and sequential topic weight vectors for sentences, and an expressive language model, capturing both short- and long-range word sequential dependencies.
For inference,we equip rGBN-RNN (decoder) with a novel recurrent variational inference network (encoder), and train it
end-to-end by maximizing an evidence lower bound (ELBO).
Different from the stacked RNN based language model in \citet{chung2017hierarchical}, which relies on three types of customized training operations (UPDATE, COPY, FLUSH) to extract multi-scale structures, the language model in rGBN-RNN learns such structures purely in a data driven manner, under the guidance of the temporally and hierarchically connected stochastic layers of rGBN. Note while both rGBN and stacked-RNN are existing methods, integrating them as a larger-context language model involves non-trivial efforts, as we need to not only carefully design how to connect the recurrent hierarchical stochastic layers of rGBN with the deterministic ones of stacked-RNN, but also design a suitable recurrent variational inference network.

The effectiveness of rGBN-RNN as a new larger-context language model is demonstrated both quantitatively, with perplexity and BLEU scores, and qualitatively, with interpretable latent structures and randomly generated sentences and paragraphs. Notably, moving beyond a usual RNN-based language model that generates individual sentences, the proposed
rGBN-RNN can generate a paragraph consisting of a sequence of semantically coherent sentences.

\section{Recurrent Hierarchical Topic-Guided Language Model}
Denote a document of $J$ sentences as $\mathcal{D}=\left(S_{1}, S_{2}, \ldots, S_{J}\right)$, %
where $S_{j}=(y_{j,1},\ldots,y_{j,T_j})$ consists of %
$T_j$ words from a vocabulary of size $V$.
Conventional statistical language models often only focus on the word sequence within a sentence. Assuming that the sentences of a document are independent of each other, they often define %
\begin{align}
\textstyle P(\mathcal{D}) &\approx \textstyle\prod_{j=1}^{J} P\left(S_{j}\right)\notag\\
&\textstyle= \prod_{j=1}^{J}\left[ p\left(y_{j,1}\right)\prod_{t=2}^{T_j} p\left(y_{j,t} \given y_{j,<t}\right)\right].
\notag
\end{align}
RNN-based neural language models define the conditional probability of each word $y_{j,t}$ given all the previous words $y_{j,<t}$ within the sentence $S_{j}$, through the softmax function of a hidden state $\hv_{j,t}$, as
\begin{align}\label{classRNN}
&p\left(y_{j,t} \given y_{j,<t}\right)= p\left(y_{j,t} \given \hv_{j,t}\right) ,\notag\\&~~~~~~\hv_{j,t} =f\left(\hv_{j,<t}, y_{j,t-1}\right),
\end{align}
{where $f(\cdot)$ is a non-linear function typically defined as an RNN cell, such as long short-term memory (LSTM) \citep{hochreiter1997long} and gated recurrent unit (GRU) \citep{Cho2014Learning}.}

{These {RNN-based} language models} are typically applied only at the {word level}, without exploiting the document context, and hence often fail to capture long-range dependencies. %
While \citet{dieng2017topicrnn}, \citet{lau2017topically}, and \citet{wang2018topic,wang2019topic_guided} remedy this issue by guiding the language model with a topic model, they still treat a document as a bag of sentences, ignoring sentence order, %
 and lack the ability to extract hierarchical and recurrent topic structures.

\begin{figure*}[!ht]
 \begin{center}
 \subfigure[]{
 \includegraphics[%
 width=4cm]{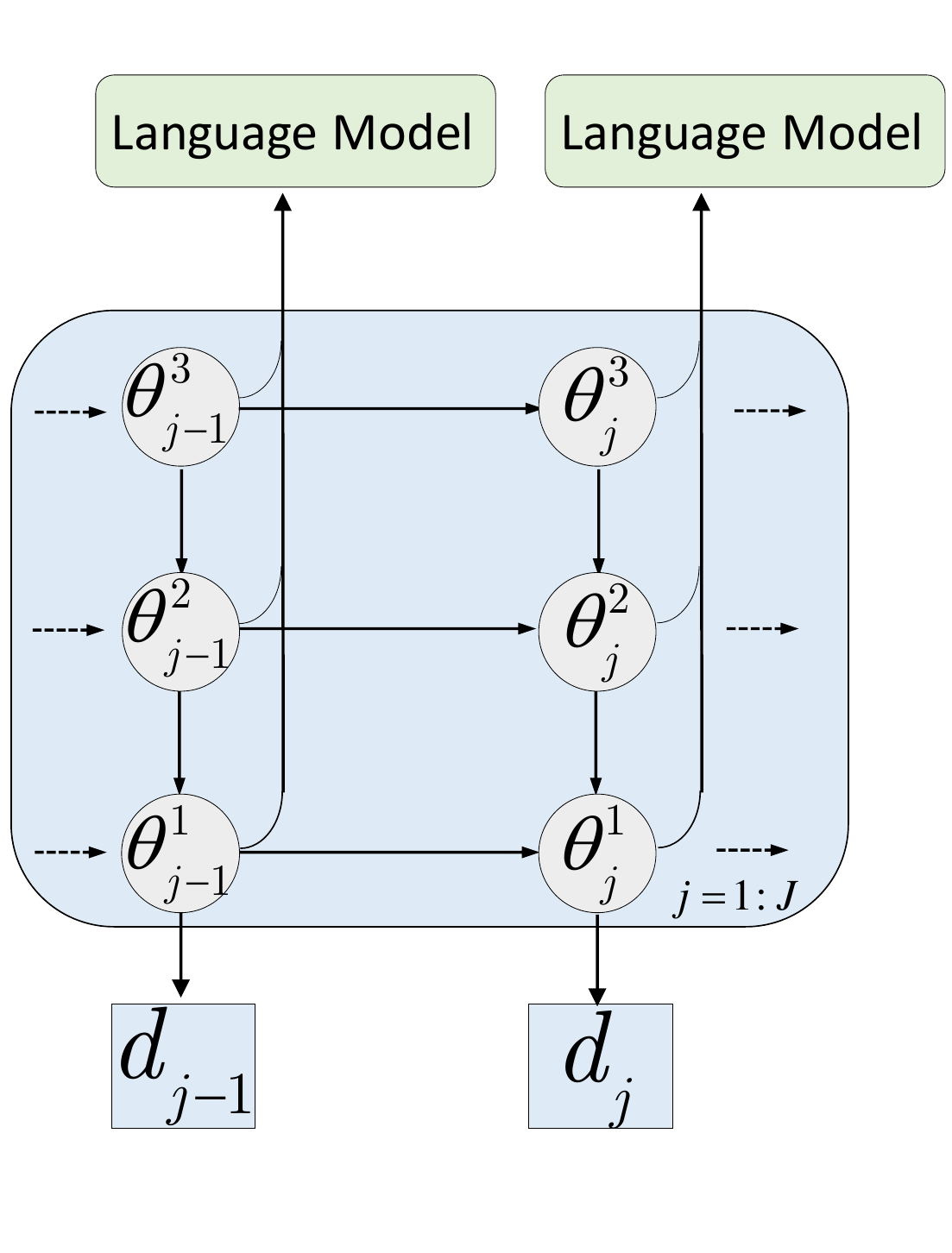}
 }\quad\quad\quad
 \subfigure[]{
 \includegraphics[%
 width=2.9cm]{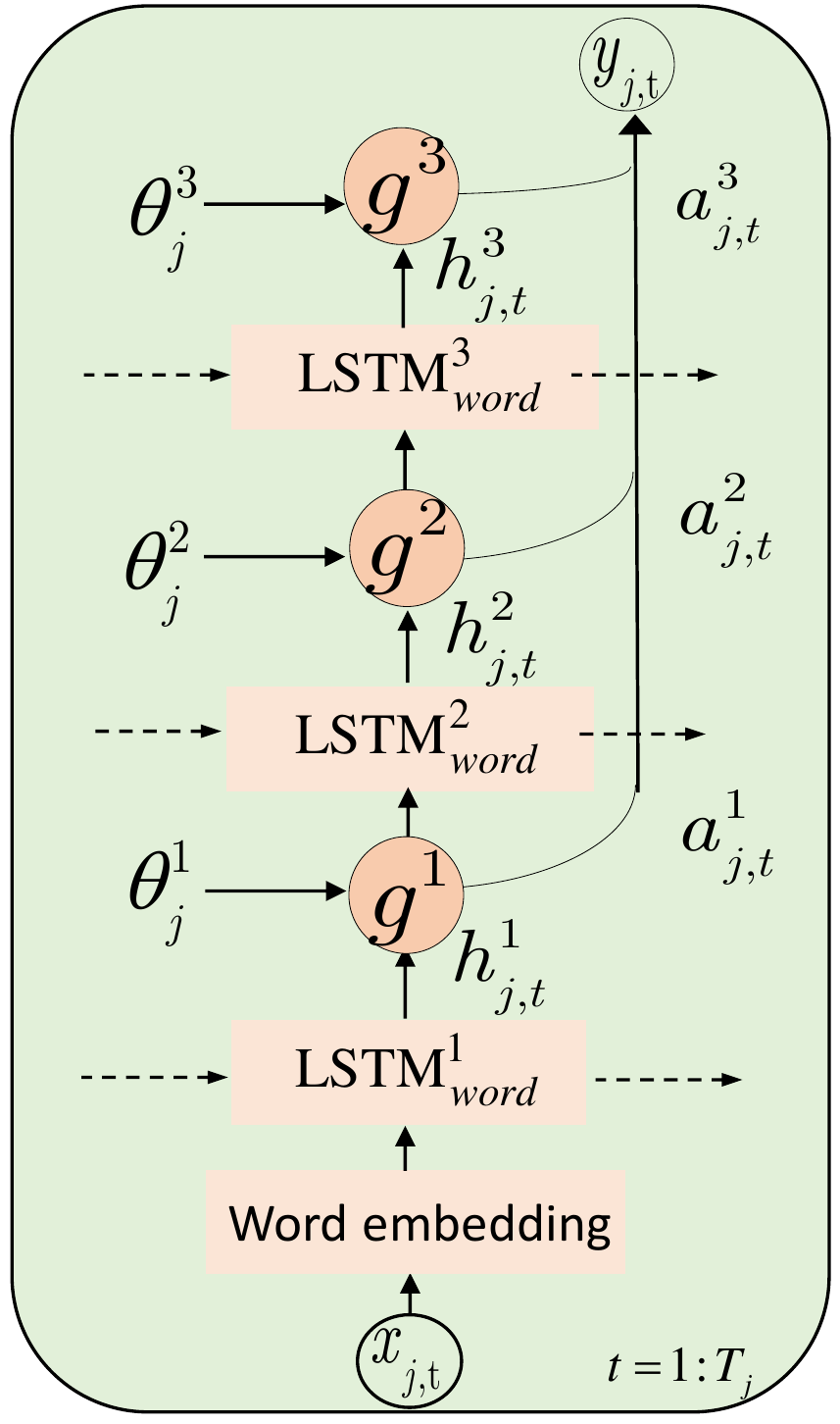}
 }\quad\quad\quad\quad
 \subfigure[]{
 \includegraphics[%
 width=5cm]{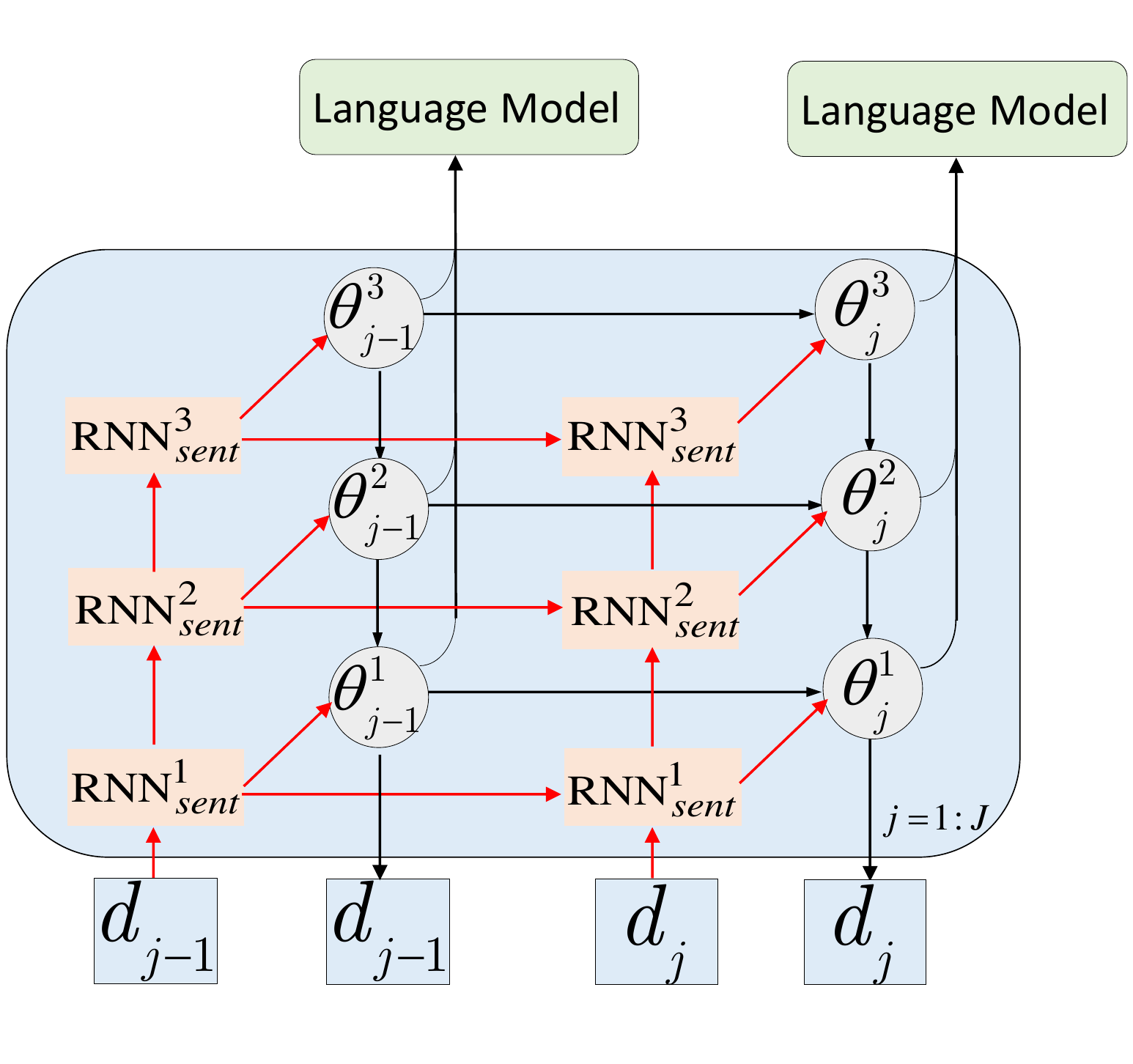}
 }\vspace{-3mm}
 \caption{ 
(a) The generative model of a three-hidden-layer rGBN-RNN, where the bottom part is the deep recurrent topic model (rGBN), document contexts of consecutive sentences are used as observed data, and upper is the language model.
(b) Overview of the language model component, where input $x_{j,t}$ denotes the $t$th word in $j$th sentence of a document, $x_{j,t}=y_{j,t-1}$, $\hv_{j,t}^{l}$ is the hidden state of the stacked RNN at time step $t$, and $\thetav_j^{l}$ is the topic weight vector of sentence $j$ at layer $l$.
(c) The overall architecture of the proposed model, including the decoder (rGBN and language model) and encoder (recurrent variational inference network), where the red arrows denote the inference of latent topic weight vectors and the black arrows denote the data generation.}
\label{fig:generative_model}
\end{center} \vspace{-3mm}
\end{figure*}

We introduce rGBN-RNN, as depicted %
 in Fig.~\ref{fig:generative_model} (a), as a new larger-context language model. It consists of two key components:
(i) a hierarchical recurrent topic model (rGBN), and (ii) a stacked RNN-based language model. We use rGBN to capture both global semantics across documents and long-range inter-sentence dependencies within a document,
and use the language model to learn the local syntactic relationships between the words within a sentence.
Similar to {\citet{lau2017topically} and \citet{wang2018topic}},
we represent a document as a sequence of sentence-context pairs as $(\{S_1,\dv_1\},\ldots,\{S_J,\dv_J\})$, {where $\dv_j\in\mathbb{Z}_+^{V_c}$ summarizes the document excluding $S_j$, specifically $(S_{1},...,S_{j-1},S_{j+1},...,S_{J})$,
into a BoW count vector, with ${V_c}$ denoting the size of the vocabulary excluding stop words.} {During testing, we redefine $\dv_j$ as the BoW vector summarizing only the preceding sentences, $i.e.,$ $S_{1:j-1}$, which will be further clarified when presenting experimental results}.
{Note a naive way to utilize sentence order is to treat each sentence as a document, use a dynamic topic model \citep{DTM} to capture the temporal dependencies of the latent topic-weight vectors, each of which is fed to the RNN to model the word sequence of its corresponding sentence. However, the sentences are often too short to be well modeled by a topic model. In our setting, as $\dv_j$ summarizes the document-level context of $S_j$, it is in general sufficiently long for topic modeling.}

\subsection{Hierarchical Recurrent Topic Model}

As shown in Fig.~\ref{fig:generative_model} (a), to model the time-varying sentence-context count vectors $\dv_j$ in document $\mathcal{D}$, the generative process of the rGBN component, from the top to bottom layers, is expressed as
\begin{gather}
 \thetav_j^{L} \sim \mbox{Gam}\left(\Pimat^{L} \thetav_{j-1}^{L} , ~\tau_0 \right),\cdots, \nonumber\\
 \thetav_j^{l} \sim \mbox{Gam}\left(\Phimat^{l+1} \thetav_j^{l+1} + \Pimat^{l} \thetav_{j-1}^{l} ,~ \tau_0 \right), \cdots, \notag\\
 \thetav_j^{1} \sim \mbox{Gam}\left(\Phimat^{2} \thetav_j^{2} + \Pimat^{1}\thetav_{j-1}^{1} , ~\tau_0 \right),\notag\\
 \dv_j \sim \mbox{Pois} \left( \Phimat^{1} \thetav_j^{1} \right), \label{DPGDS}%
\end{gather}
 where $\thetav_j^{l} \in \mathbb{R}_+^{K_l}$ denotes the gamma distributed topic weight vector of sentence $j$ at layer~$l\in\{1,\ldots,L\}$, $\Pimat^{l}\in \mathbb{R}_{+}^{K_{l} \times K_{l}}$ the transition matrix of layer~$l$ that captures cross-topic temporal dependencies, $\Phimat^{l}\in \mathbb{R}_{+}^{K_{l-1} \times K_{l}}$ the loading matrix at layer~$l$, $K_l$ the number of topics of layer $l$, and $\tau_0 \in \mathbb{R}_+$ a scaling hyperparameter. %
At $j=1$, $\thetav_1^{l} \sim \mbox{Gam}\left(\Phimat^{l+1} \thetav_1^{l+1} , \tau_0 \right)$ for $l=1,\ldots,L-1$ and
$\thetav_1^{L} \sim \mbox{Gam}\left(\nu , \tau_0 \right)$, where $\nu = \mathbf{1}_{K_L}$.
Following {\citet{guo2018deep} and \citet{zhou2015poisson}}, the Dirichlet priors are placed on the columns of $\Pimat^{l}$ and $\Phimat^{l}$, $i.e.$, $\piv_k^{l}$ and $\phiv_k^{l}$ ,
which not only makes the latent representation more identifiable and interpretable, but also facilitates inference.
The count vector $\dv_j$ can be factorized into the product of $\Phimat^{1}$ and $\thetav_j^{1}$ under the Poisson likelihood.
The shape parameters of $\thetav_j^{l} \in \mathbb{R}_+^{K_l}$ can be
factorized into the sum of $\Phimat^{l+1} \thetav_j^{l+1}$, capturing inter-layer hierarchical dependence, %
and $\Pimat^{l} \thetav_{j-1}^{l}$,
capturing intra-layer temporal dependence. %
rGBN not only captures the document-level word occurrence patterns inside the training text corpus, but also the sequential dependencies of the sentences inside a document.
Note ignoring the recurrent structure, rGBN {will} reduce to the gamma belief network (GBN) of \citet{GBN}, which can be reformulated as a multi-stochastic-layer deep generalization of LDA \citep{cong2017deep}; if setting the number of stochastic hidden layer as $L=1$, GBN reduces to Poisson factor analysis \citep{zhou2012beta,NBP2012} .
If ignoring its hierarchical structure ($i.e.$, $L=1$), rGBN reduces to Poisson--gamma dynamical systems of \citet{ScheinWallachZhou_PGDS_2016} that generalizes the gamma Markov chain of \citet{GP-DPFA2015} by adding latent state transitions. %
We refer to the rGBN-RNN without its recurrent structure as GBN-RNN, which no longer models sequential sentence dependencies;
see Appendix \ref{sec:GBN-RNN} for more details.

\subsection{{Language Model}}
Different from a conventional {RNN-based} language model, which predicts the next word only using the preceding words within the sentence, we integrate the hierarchical recurrent topic weight vectors $\thetav_j^{l}$ into the language model to predict the word sequence in the $j$th sentence.
Our proposed language model is built upon the stacked RNN proposed in {\citet{graves2013generating} and \citet{chung2017hierarchical}}, but with the help of rGBN, it no longer requires specialized training heuristics to extract multi-scale latent structures.
As shown in Fig. \ref{fig:generative_model} (b), to generate $y_{j,t}$, the $t^{\text{th}}$ token of sentence $j$ in a document, we construct the hidden states $\hv_{j,t}^{l}$ of the language model, from the bottom to top layers, as
\begin{equation}\label{RNN_hiddenstate}
\hv_{j,t}^{l}=\left\{\begin{array}{ll}{\mathrm{LSTM}_{\mathrm{word}}^{l}\left(\hv_{j, t-1}^{l}, \boldsymbol{W}_{\boldsymbol{e}}\left[x_{j, t}\right]\right)} ,& {\text {if } l=1 } ,\\
\mathrm{LSTM}_{\mathrm{word}}^{l}\left(\hv_{j, t-1}^{l},\av_{j, t}^{{l-1}} \right), & {\text {if } 1<l\le L } ,
\end{array}\right.
\end{equation}
where $\mathrm{LSTM}_{\mathrm{word}}^{l}$ denotes the word-level LSTM at layer~$l$, ${\boldsymbol{W}_{\boldsymbol{e}}}%
$ are word embeddings
to be learned, and $x_{j,t}=y_{j,t-1}$. Note $\av_{j, t}^{l}$ denotes the coupling vector, which combines the temporal topic weight vectors $\thetav_j^{l}$ and hidden output of the word-level LSTM $\hv_{j, {t}}^{l}$ at each time step $t$. Following %
 \citet{lau2017topically}, we realize $\av_{j, t}^{l}= g^{l}\left(\hv_{j, t}^{l}, \thetav_j^{l} \right)$ with a gating unit similar to a GRU \citep{Cho2014Learning}, described as
\begin{equation}
    \av_{j, t}^{l} =\left(1-\zv_{j,t}^l\right) \odot \hv_{j, t}^{l}+\zv_{j,t}^l \odot \hat{\hv}_{j,t}^l,\label{coupling_vector}
\end{equation}
where
\begin{equation}\notag %
\begin{aligned}
\small
\zv_{j,t}^l &=\sigma\left(\mathbf{W}_{z}^l \thetav_j^{l} +\mathbf{U}_{z}^l \hv_{j, t}^{l}+\mathbf{b}_{z}^l\right), \\
\rrv_{j,t}^l &=\sigma\left(\mathbf{W}_{r}^l \thetav_j^{l}+\mathbf{U}_{r}^l \hv_{j, t}^{l}+\mathbf{b}_{r}^l\right), \\
\hat{\hv}_{j,t}^l &=\tanh \left(\mathbf{W}_{h}^l \thetav_j^{l}+\mathbf{U}_{h}^l\left(\rrv_{j,t}^l \odot \hv_{j, t}^{l}\right)+\mathbf{b}_{h}^l\right).
\end{aligned}
\end{equation}

 Denote $\av_{j,t}^{1:L}$ as the concatenation of $\av_{j,t}^{l}$ across all layers and $\Wmat_o$ as a weight matrix with $V$ rows;
different from \eqref{classRNN}, the conditional probability of $y_{j, t}$ becomes
\begin{equation}
 p\left(y_{j, t} \given y_{j,<t}, {\thetav_{j}^{1:L}} \right) =\operatorname{softmax}\left(\Wmat_o \av_{j, t}^{1:L}\right).
 \end{equation}
{There are two main reasons for combining all the latent representations $\av_{j, t}^{1:L}$ for language modeling. First, the latent representations exhibit different statistical properties at different stochastic layers of rGBN-RNN, and hence are combined together to enhance their representation power. Second, having ``skip connections'' from all hidden layers to the output one makes it easier
to train the proposed network, reducing the number of processing steps between the bottom of the network and the top and hence mitigating the ``vanishing gradient'' problem \citep{graves2013generating}.}

{To sum up, as depicted in Fig. \ref{fig:generative_model} (a), the topic weight vector $\thetav_j^{l}$ of sentence $j$ quantifies the topic usage of its document context $\dv_j$ at layer $l$. %
 It is further used as an additional feature of the language model to guide the word generation %
 inside sentence $j$, as shown in Fig. \ref{fig:generative_model} (b). It is clear that rGBN-RNN has two temporal structures: {a deep recurrent topic model} to extract the temporal topic weight vectors from the sequential document contexts, and a {language model} to estimate the probability of each sentence given its corresponding hierarchical topic weight vector.
Characterizing the word-sentence-document hierarchy to incorporate %
both {intra- and inter-sentence} information, rGBN-RNN learns more coherent and interpretable topics and increases the generative power of the language model.
Distinct from existing topic-guided language models, the temporally related hierarchical topics of rGBN exhibit different statistical properties across layers, which helps better guide {language model} to improve its language generation ability.}

\subsection{Model Likelihood and Inference}

 For rGBN-RNN, given $\{\Phimat^{l},\Pimat^{l}\}_{l=1}^{L} $, the marginal likelihood of the sequence of sentence-context pairs $(\{\sv_{1},\dv_1\},\ldots,\{\sv_{J},\dv_J\})$ of document $ \mathcal{D}$ is defined as
\begin{align}\label{likelihood2}
&\resizebox{0.8\hsize}{!}{$ P\left( \mathcal{D} %
\given \{\Phimat^{l},\Pimat^{l}\}_{l=1}^{L} \right)=
 \int \textstyle\prod_{j = 1}^{J} p\left(\dv_{j}\given \Phimat^{1}\thetav_j^{1}\right)$} \notag \\
&\resizebox{0.89\hsize}{!}{$ \left[\prod_{t = 1}^{T_j}p\left(y_{j,t}\given y_{j,<t},\thetav_j^{1:L}\right)\right]
\left[\prod_{l = 1}^{L} p\left(\thetav_{j}^{l}\given
\ev_j^l
,\tau_0 \right) \right]d{\thetav_{1:J}^{1:L}}$},
\end{align}
where $\ev_j^{l}: = \Phimat^{l+1}\thetav_j^{l+1}+\Pimat^{l}\thetav_{j-1}^{l}$. %
The inference task is to learn the parameters of both the topic model and language model components.
One naive solution is to alternate the training between these two components in each iteration: First,
the topic model is trained %
using %
a sampling based iterative algorithm
provided in \citet{guo2018deep}; Second, the language model %
is trained with
 maximum likelihood estimation under a standard cross-entropy loss. While this naive %
 solution can utilize readily available %
 inference algorithms for both rGBN and {the language model}, it may suffer from stability and convergence issues. Moreover, the need to perform a sampling based iterative algorithm for rGBN inside each iteration %
 limits the scalability of the model for both training and testing.
To this end,
we introduce a recurrent variational inference network (encoder) to learn the latent temporal topic weight vectors {$\thetav_{1:J}^{1:L}$}.
{Denoting $Q=\prod_{j=1}^J \prod_{l=1}^L q(\thetav_j^{l}\given {\dv_{j}})$}, an ELBO of the log marginal likelihood shown in \eqref{likelihood2}
can be constructed as
 \begin{align}\label{ELBO-of-our model2}
\small
&\resizebox{0.99\hsize}{!}{$ L = \sum_{j=1}^J {\sum_{l=1}^L} \mathbb{E}_Q\left[ \ln p\left( \dv_j \given \, \Phimat^{1}\thetav_j^{1} \right)+ \sum_{t=1}^{T_j} \ln p\left( {y_{j,t}}\given y_{j,<t},\thetav_j^{1:L}\right)\right]$} \notag \\
&~~~~\resizebox{0.6\hsize}{!}{$- \sum_{j=1}^J \sum_{l=1}^L \mathbb{E}_Q \left[ \ln \frac{q\left( \thetav_j^{l}\given \dv_{\le j} \right) }{p \left( \thetav_j^{l} \given
{\ev_j^{l}},\tau_0\right)} \right],$}
\end{align}
which unites both the terms primarily responsible for training the recurrent hierarchical topic model component, and terms for training the RNN language model component.
Similar to \citet{Zhang2018WHAI}, {we define
$q(\thetav_j^{l} \,\given {\dv_{j}}) = \mbox{Weibull}(\kv_j^{l},\lambdav_j^{l})$}, a random sample from which can be obtained by transforming standard uniform noises $\epsilonv_{j}^{l}$ as %
\begin{equation}\label{update_theta}\textstyle
\small
\thetav_{j}^{l} = {\lambdav_{j}^{l}} \big(-\ln(1-{\epsilonv_{j}^{l}})\big) ^ {1/\kv_{j}^{l}}.
\end{equation}
To capture the temporal dependencies between the topic weight vectors, %
both $\kv_j^{l}$ and $\lambdav_j^{l}$,
{from the bottom to top layers, can be expressed as}
\begin{align}\label{update_network_parameters}
\hv_j^{s,l} = \mathrm{RNN}_{\mathrm{sent}}^{l}\big(\hv_{j-1}^{s,l}, \hv_{j}^{s,l-1}\big),\notag\\
\kv_j^{l} = f_{\kv}^{l}\big(\hv_j^{s,l}\big),~~
\lambdav_j^{l} = f_{\lambdav}^{l}\big(\hv_j^{s,l}\big),
\end{align}
where $\hv_j^{s,0}=\dv_j$, $\hv_0^{s,l}=0$, {$\mathrm{RNN}_{\mathrm{sent}}^{l}$ denotes the
sentence-level recurrent encoder at layer $l$ implemented with a basic RNN cell, {capturing the sequential relationship between sentences within a document}, $\hv_j^{s,l}$ denotes the hidden state of $\mathrm{RNN}_{\mathrm{sent}}^{l}$, and superscript $s$ in $\hv_j^{s,l}$ denotes ``sentence-level RNN'' used to distinguish the hidden state of language model in \eqref{RNN_hiddenstate}
}.
Note both $f_{\kv}^{l}$ and $f_{\lambdav}^{l}$ are nonlinear functions mapping state $\hv_{j}^{s,l}$ to the parameters of $\thetav_j^{l}$, implemented with $f(\xv) = \textrm{ln}(1+\textrm{exp}(\Wmat\xv+\bv))$.

Rather than finding a point estimate of the global parameters $\{\Phimat^{l},\Pimat^{l}\}_{l=1}^{L} $ of the rGBN,
we adopt a hybrid inference algorithm by combining TLASGR-MCMC described in {\citet{cong2017deep} and \citet{Zhang2018WHAI}} and our proposed recurrent variational inference network. In other words, the global parameters $\{ \Phimat^{l},\Pimat^{l} \}_{l=1}^{L}$ can be sampled with TLASGR-MCMC, while the parameters of the {language model} and recurrent variational inference network, denoted by $\Omegamat$, can be updated via stochastic gradient descent (SGD) by maximizing the ELBO in \eqref{ELBO-of-our model2}.
We describe a hybrid variational/sampling inference for rGBN-RNN in Algorithm~\ref{Algorithm} and provide more
details about sampling $\{ \Phimat^{l},\Pimat^{l} \}_{l=1}^{L}$ with TLASGR-MCMC in Appendix \ref{sec:SGMCMC for GBN-RNN}. {We defer the details on model complexity to Appendix \ref{sec:parameters}.}

To sum up, as shown in Fig. \ref{fig:generative_model} (c), the proposed rGBN-RNN works with a recurrent variational autoencoder inference framework, which takes the document context of the $j$th sentence within a document as input and learns hierarchical topic weight vectors $ \thetav_j^{1:L}$ that evolve sequentially with $j$. The learned topic vectors in different layer are then used to reconstruct the document context input and as an additional feature for the {language model} to generate the $j$th sentence.

\begin{algorithm}[!t]
\footnotesize
\caption{ Hybrid TLASGR-MCMC and recurrent autoencoding variational inference for rGBN-RNN.}
\begin{algorithmic}
 \STATE Set mini-batch size $m$ and the number of layer $L$
 \STATE Initialize encoder and neural language model {parameters} $\Omegamat$, and topic model {parameters} {$ \{ \Phimat^{l},\Pimat^{l} \}_{l=1}^{L}$}.

 \FOR{$iter = 1,2, \cdots$ }
 \STATE %
 Randomly select a mini-batch of $m$ documents consisting of $J$ sentences to form a subset $\Xmat = \{ \dv_{i,{1:J}},\sv_{i,{1:J}} \}_{i = 1}^{m}$;

 Draw random noise $\left\{ {{\epsilonv _{i,j}^{l}}} \right\}_{i = 1,j=1,l=1}^{m,J,L}$ from uniform distribution;

 Calculate $ {\nabla _{\Omegamat }} L\left( \Omegamat,\Phimat ^{l},\Pimat ^{l} ;{\Xmat},{\epsilonv_{i,j}^{l}} \right)$ according to \eqref{ELBO-of-our model2}, and update $\Omegamat$;
 Sample $\thetav _{i,j}^{l}$ from \textbf{\eqref{update_theta}} and \eqref{update_network_parameters} via $\Omegamat$ to update $ \{ \Pimat^{l} \}_{l=1}^{L}$ and $ \{ \Phimat^{l} \}_{l=1}^{L}$, as described in Appendix \ref{sec:SGMCMC for GBN-RNN};
 \ENDFOR
\end{algorithmic}\label{Algorithm}
\end{algorithm}

\section{Experimental Results}

We consider three publicly available corpora, including APNEWS, IMDB, and BNC. The links, preprocessing steps, and summary statistics for them are deferred to Appendix \ref{sec:data}. We consider a {recurrent variational inference network} for rGBN-RNN to infer ${\thetav_j^{l}}$, as shown in Fig.~\ref{fig:generative_model} (c), whose number of hidden units in \eqref{update_network_parameters} are set the same as the number of topics at the corresponding layer. %
Following \citet{lau2017topically}, word embeddings are pre-trained 300-dimension word2vec Google News vectors (\href{https://code.google.com/archive/p/word2vec/}{https://code.google.com/archive/p/word2vec/}).
Dropout with a rate of $0.4$ is used to the input of the {stacked-RNN} at each layer, $i.e.$, ${{\av}_{j,t}^{l}}$ or $\boldsymbol{W}_{\boldsymbol{e}}\left[x_{j, t}\right]$ in \eqref{RNN_hiddenstate}. %
The gradients are clipped if the norm of the parameter vector exceeds $5$. We use the Adam optimizer \citep{kingma2015adam} with learning rate $10^{-3}$.
The length of an input sentence is fixed to $30$.
We set the mini-batch size as $8$, number of training epochs as $5$, and %
$\tau_0$~as~$1$. Python (TensorFlow) code is provided at \rr{\url{https://github.com/Dan123dan/rGBN-RNN}}

\subsection{Quantitative Comparison}
\textbf{Perplexity:}
For fair comparison, we use standard language model perplexity as the evaluation metric. We consider the following baselines:
\textit{1)} A standard LSTM language model \citep{hochreiter1997long}; \textit{2)} LCLM \citep{Tian2016Larger}, a larger-context language model that incorporates context from preceding sentences, %
{which are treated} as a bag of words; \textit{3)}
A standard LSTM language model incorporating the topic information of a separately trained LDA (LDA+LSTM); \textit{4)} Topic-RNN
\citep{dieng2017topicrnn}, a hybrid model rescoring the prediction of the next word by incorporating the topic information through a linear transformation; \textit{5)} TDLM \citep{lau2017topically}, a joint learning framework that learns a convolution based topic model and a language model simultaneously;
\textit{6)} TCNLM \citep{wang2018topic}, which extracts the global semantic coherence
of a document via a neural topic model, with the probability of each learned latent
topic further adopted to build a mixture-of-experts language model; \textit{7)} TGVAE~\citep{wang2019topic_guided}, combining a variational auto-encoder based neural sequence model with a neural topic model; \textit{8)} GBN-RNN, a simplified rGBN-RNN that removes the recurrent structure of its rGBN component;
\textit{9)} rGBN-RNN-flipped, which is an additional architectural variation of the proposed rGBN-RNN that modifies $\thetav_j^3$ and $\thetav_j^1$ shown in Fig.~\ref{fig:generative_model}(b) by swapping their locations;
\textit{10)} Transformer-XL \citep{dai2019transformerxl}, which enables learning dependency beyond a fixed length by introducing a recurrence mechanism and a novel position encoding scheme into the Transformer architecture;
\textit{11)} GPT-2 \citep{radford2019language}, which can be realized by a generative pre-training of a Transformer-based language model on a diverse set of unlabeled text, followed by discriminative {fine}-tuning on each specific dataset.

For rGBN-RNN, to ensure the information about the words in the $j$th sentence to be predicted is not leaking through the {sequential} document context vectors at the testing stage, {the input $\dv_j$ in~\eqref{update_network_parameters}} only summarizes the {preceding sentences $S_{<j}$}.
For GBN-RNN, following TDLM \citep{lau2017topically} and TCNLM \citep{wang2018topic}, all the sentences in a document, excluding the one being predicted, are used to obtain the BoW document context.
As shown in Table~\ref{Tab:perplexity and topic coherence}, rGBN-RNN outperforms all {RNN-based} baselines, and the trend of improvement continues as its number of layers increases, indicating the effectiveness of incorporating recurrent hierarchical topic information into language generation.
rGBN-RNN consistently outperforms GBN-RNN, suggesting the benefits of exploiting the sequential dependencies of the sentence-contexts
for language modeling.

In Table \ref{Tab:perplexity and topic coherence}, we 
further compare the number of parameters between %
various %
language models, %
where we follow the convention to ignore the word embedding layers.
The number of parameters for some models are not reported, 
as we could not find sufficient information from their corresponding papers or code to provide accurate estimations. 
When used for language generation at the testing stage, rGBN-RNN no longer needs its topics $\{\Phimat^{l}\}$, whose parameters are hence not counted. Note the number of parameters of the topic model component is often dominated by that of the language model component. { Table~\ref{Tab:perplexity and topic coherence} {suggests rGBN-RNN, with its hierarchical and temporal topical guidance, achieves better performance with fewer parameters than comparable RNN-based language models.}}

Note that for language modeling, there has been significant recent interest in replacing RNNs with the Transformer \citep{vaswani2017attention}, which consists of stacked multi-head attention modules, and its variants \citep{dai2019transformerxl,devlin2019bert,radford2018improving,radford2019language}.
{For comparison, we also report the performance of GPT-2 and Transformer-XL, two Transformer-based models.
Although shown in Table \ref{Tab:perplexity and topic coherence}, GPT-2
can obtain better performance than our proposed models, %
GPT-2
has significantly more parameters and requires a huge text corpus for pre-training.}
{For example, GPT-2 with 12L \citep{radford2019language}
has 117M parameters, while the proposed rGBN-RNN with three hidden layers has as few as 7.3M parameters for language modeling.
Moreover, without pre-training, we have tried training the GPT-2 directly with the APNEWS corpus {on one machine with 4 NVIDIA RTX 2080 Ti GPUs}: even after running 24 hours, the perplexity stays above 600 and does not show a clear trend of improvement as the time progresses.
Therefore, we only display in Fig.~\ref{fig:time_of_models} how Transformer-XL and rGBN-RNN %
behave during training, by showing
the test perplexity of APNEWS documents.
It is clear that rGBN-RNN is able to fit the data well, while Transformer-XL behaves well during the early stage of training, it shows a clear trend of overfitting as the training progresses further, possibly because it has an overly large number of model parameters, making it prone to overfitting and hence difficult to generalize. %

\begin{figure}[!t]
\begin{center}
\includegraphics[width=.3\textwidth]{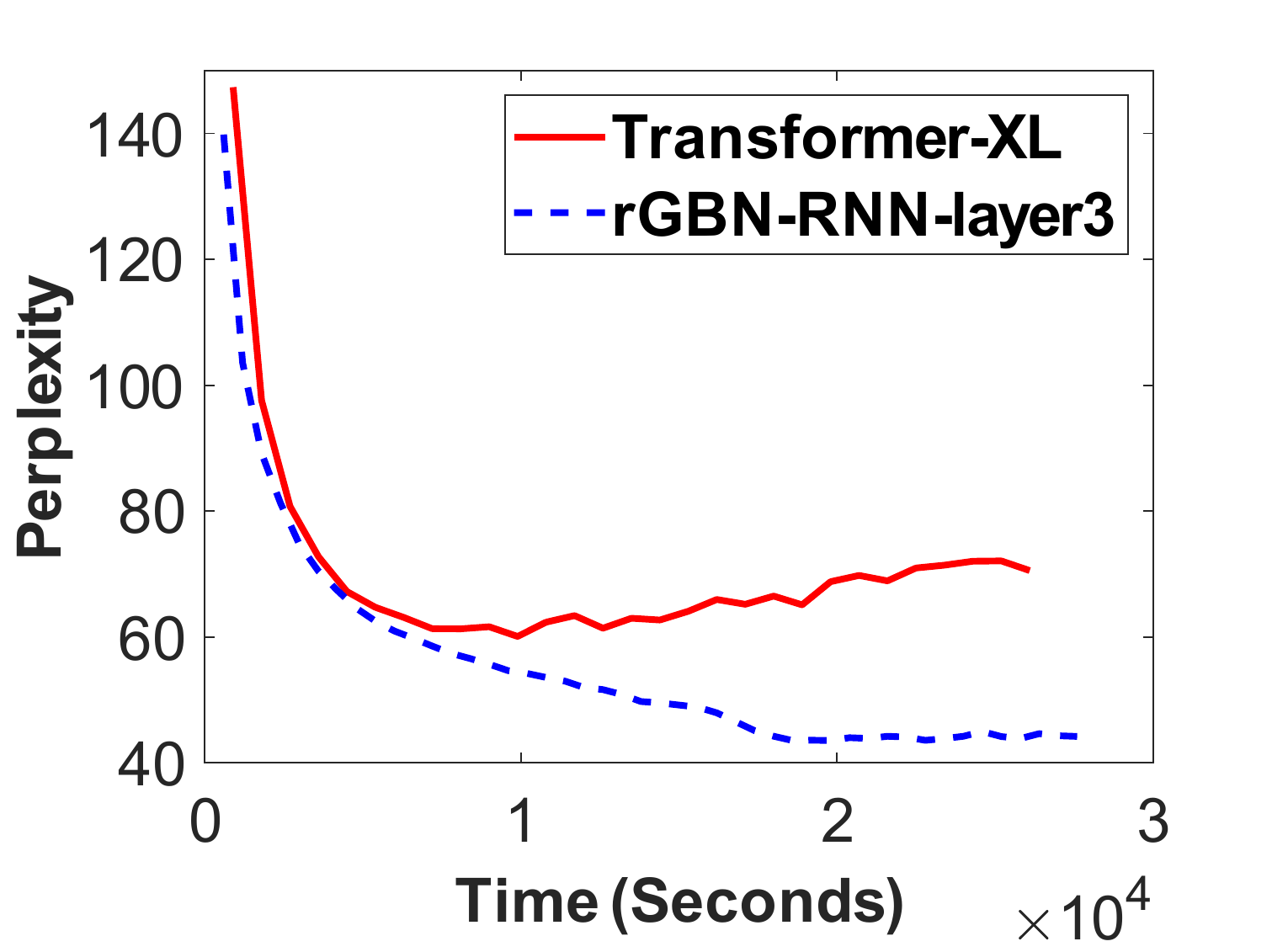}
\vspace{-3mm}
\caption{ 
Comparison of Transformer-XL and rGBN-RNN on the test perplexity as a function of
training time on APNEWS. }
\label{fig:time_of_models} \vspace{-4mm} %
\end{center}
\end{figure}

\begin{table*}[th]
\centering
\small
\caption{ Comparison of perplexity on three different datasets and the number of parameters when used for language generation. }
\resizebox{.98\textwidth}{!}{
\begin{tabular}{cccccc|ccc}
\toprule
\mr{2}*{Model} & \mr{2}*{LSTM Size} & \mr{2}*{\#LM Param} & \mr{2}*{Topic Size} & \mr{2}*{\#TM Param} & \mr{2}*{\#All Param} & \mc{3}{c}{Perplexity} \\ \cline{7-9}
&& &&& & APNEWS & IMDB & BNC \\ \midrule
\mr{2}*{LCLM \citep{Tian2016Larger}} & 600 & --- & --- & --- & ---& 54.18 & 67.78 & 96.50 \\
& 900-900 & --- & --- & --- & ---& 50.63 & 67.86 & 87.77 \\ \midrule
\mr{2}*{LDA+LSTM} & 600 & 2.16M & 100 & 0M & 2.16M& 55.52 & 69.64 & 96.50 \\
& 900-900 & 9.72M & 100 & 0M & 9.72M& 50.75 & 63.04 & 87.77 \\\midrule
\mr{2}*{TopicRNN \citep{dieng2017topicrnn}} & 600 & 4M & 100 & 4M& 4M & 54.54 & 67.83 & 93.57 \\
& 900-900 & 4M & 100 & 4M & 4M & 50.24 & 61.59 & 84.62 \\ \midrule
\mr{2}*{TDLM \citep{lau2017topically}} & 600 & 3.33M & 100 & 0.019M & 3.35M & 52.75 & 63.45 & 85.99 \\
& 900-900 & 13.36M & 100 & 0.019M & 13.38M & 48.97 & 59.04 & 81.83 \\ \midrule
\mr{2}*{TCNLM \citep{wang2018topic}} & 600 & --- & 100 & --- & --- & 52.63 & 62.64 & 86.44 \\
& 900-900 & --- & 100 & --- & ---& 47.81 & 56.38 & 80.14 \\ \midrule
TGVAE \citep{wang2019topic_guided} & 600 & --- & 50 & --- & ---& 48.73 & 57.11 & 87.86 \\ \midrule
\mr{3}*{basic-LSTM \citep{hochreiter1997long} } & 600 & 2.16M & --- & --- & 2.16M & 64.13& 72.14 & 102.89 \\
& 900-900 & 10.80M & --- & --- & 10.80M & 58.89 & 66.47 & 94.23 \\
& 900-900-900 & 17.28M & --- & --- & 17.28M & 60.13 & 65.16 & 95.73 \\ \midrule
\mr{3}*{GBN-RNN} & 600 & 3.4M & 100 & 0.02M & 3.42M & 47.42& 57.01 & 86.39 \\
& 600-512 & 6.5M & 100-80 & 0.04M & 6.54M & 44.64 & 55.42 & 82.95 \\
& 600-512-256 & 7.2M & 100-80-50 & 0.05M & 7.25M & 44.35 & 54.53 & 80.25 \\ \midrule
\mr{3}*{rGBN-RNN} & 600 & 3.4M & 100 & 0.03M & 3.43M & 46.35 &55.76 & 81.94 \\
& 600-512 & 6.5M & 100-80 & 0.06M & 6.56M & 43.26 & 53.82 & 80.25 \\
& 600-512-256 & 7.2M & 100-80-50 & 0.07M & 7.27M & \textbf{42.71} & \textbf{51.36} & \textbf{79.13} \\ \midrule
{rGBN-RNN-flipped} & 600-512-256&7.2M& 100-80-50&0.07M&7.27M & 43.55 & 53.28 & 81.12 \\ \midrule  \midrule \
 Transformer-XL \citep{dai2019transformerxl} &--- & 151M &---&---&151M & 58.73 & 60.11 & 97.14 \\ \midrule
{Pretrained GPT-2}  \citep{radford2019language} &--- & 117M &--- &---&117M&
\textbf{35.78} & \textbf{44.71} & \textbf{46.04} \\ \bottomrule
\end{tabular}\label{Tab:perplexity and topic coherence}}\vspace{-5.5mm}
\end{table*}

\begin{figure*}[!ht]
\setlength{\abovecaptionskip}{0.cm}
\setlength{\belowcaptionskip}{-0.cm}
 \begin{center}
 \subfigure[BLEU-3]{
 \includegraphics[width=0.35\textwidth]{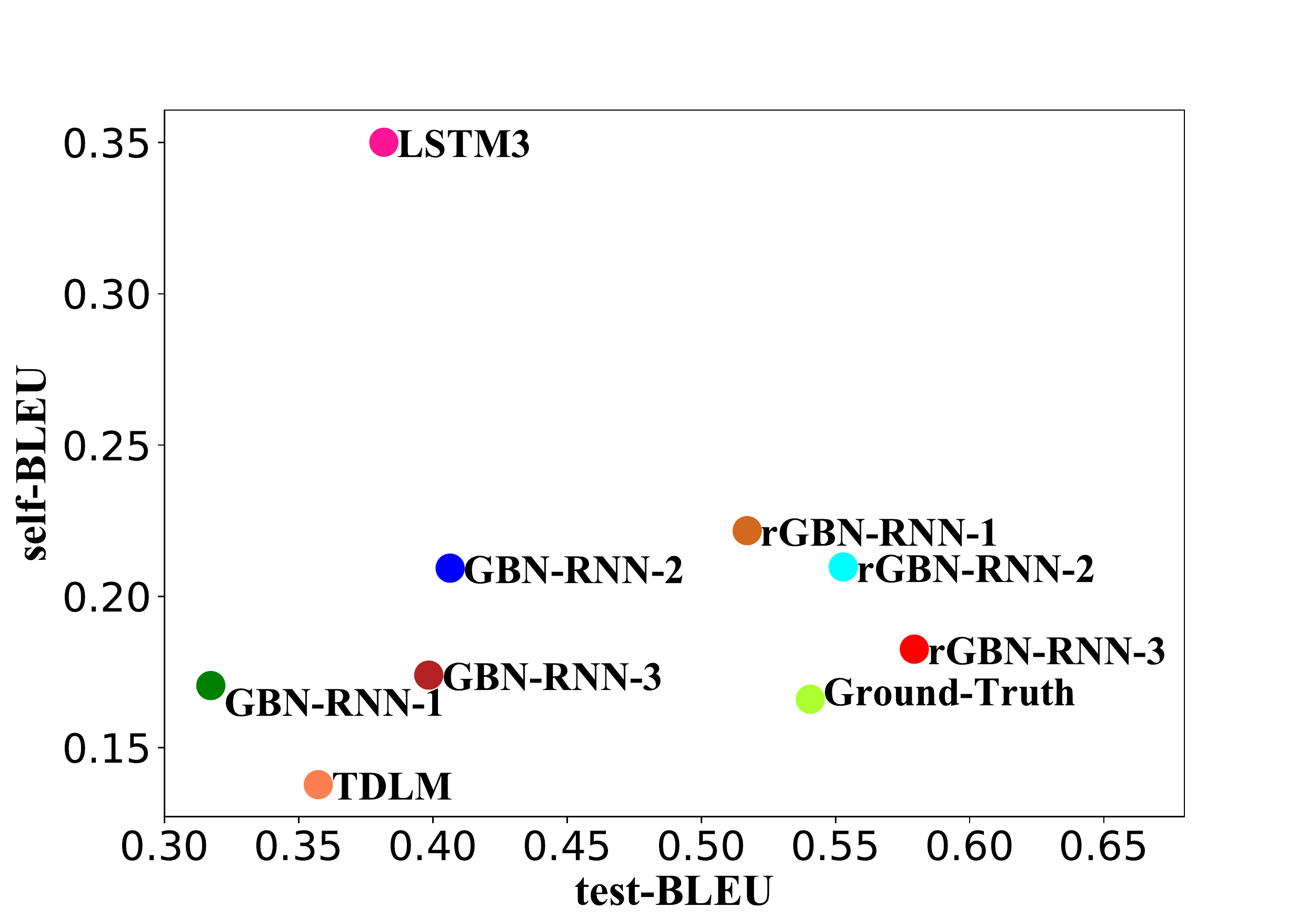}
 }
 \subfigure[BLEU-4]{
 \includegraphics[width=0.35\textwidth]{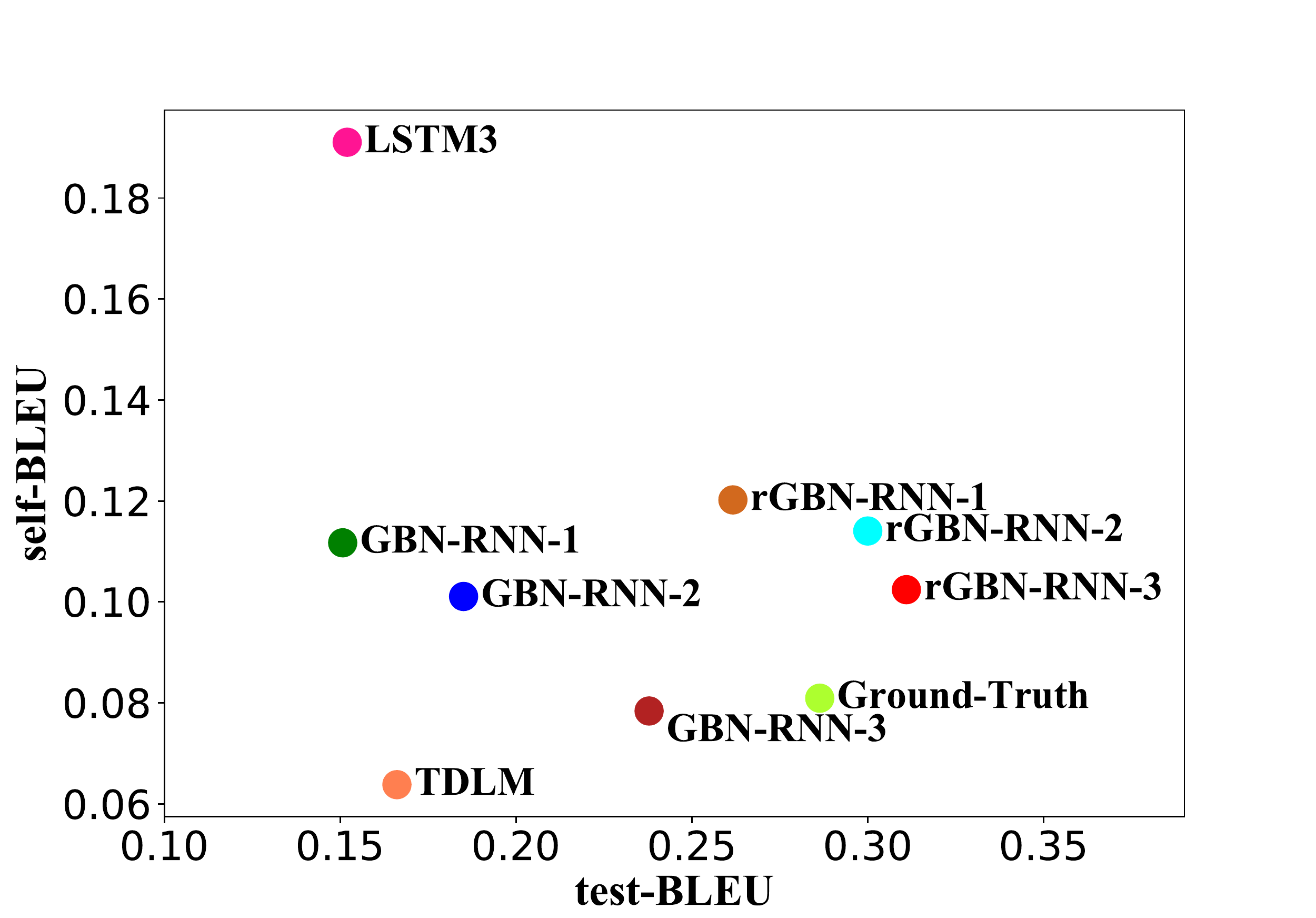}
 }\vspace{-3mm}
 \caption{ 
 {BLEU scores of different methods for BNC. BLEU scores towards the lower right corner are preferred.}
 }
 \label{fig:BLEU} \vspace{-1.mm}
\end{center}

\end{figure*}

\begin{figure*}[!ht]
\begin{center}
\includegraphics[width=.9\textwidth]{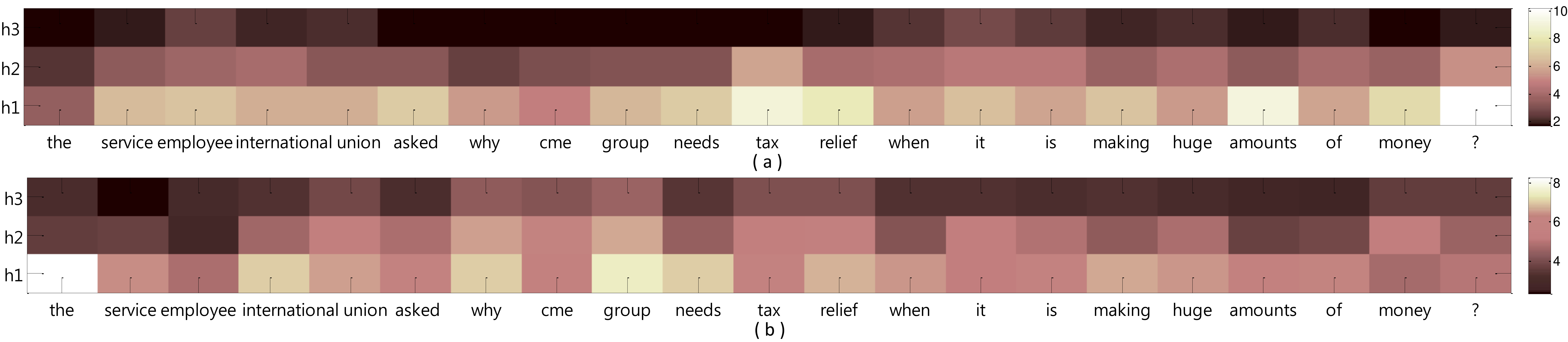}
\vspace{-3mm}
\caption{ 
{Visualizing the $L_2$ norms of the hidden states of rGBN-RNN and GBN-RNN, shown in the top and bottom rows, respectively. }}
\label{fig:hidden_states} \vspace{-4mm} %
\end{center}
\end{figure*}

\begin{figure*}[!ht]
\begin{center}
\includegraphics[height=10.5cm
]{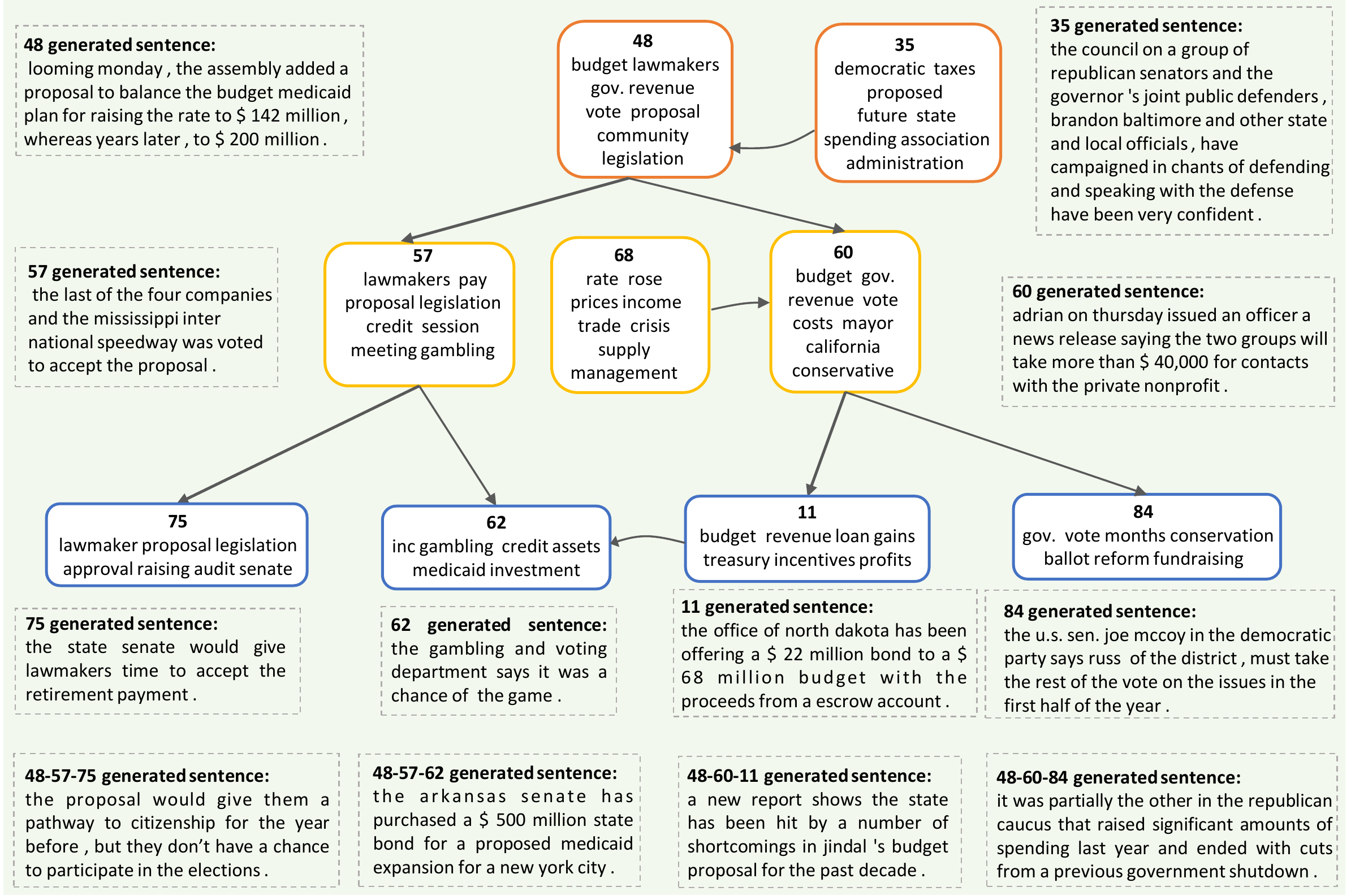}\vspace{-3mm}
\caption{\footnotesize{
Example topics and their hierarchical and temporal connections %
inferred by a three-hidden-layer rGBN-RNN from the APNEWS corpus, and the generated sentences under topic guidance.
{Top words of each topic at layers 3, 2, and 1 are shown in orange, yellow, and blue boxes, respectively,} and each sentence is shown in a dotted line box labeled with the corresponding topic index. Sentences generated with a combination of topics at different layers are shown at the bottom. See the Appendix for analogous plots for both IMDB and BNC.}
}
\label{fig:sentence_topic} \vspace{-2mm} %
\end{center}
\end{figure*}

From a structural point of view, we consider the proposed rGBN-RNN as complementary to rather than competing with Transformer based language models, and consider replacing RNN with Transformer to construct a GBN or rGBN guided Transformer as a promising future extension.
}

\textbf{BLEU:} %
Following \citet{wang2019topic_guided}, we use test-BLEU to evaluate the quality of generated sentences with a set of real test sentences as the reference, and self-BLEU to evaluate the diversity of the generated sentences \citep{zhu2018texygen}.
Given the global parameters of the deep recurrent topic model (rGBN) and language model, we can generate the sentences by following the data generation process of rGBN-RNN: we first generate {topic weight vector $\thetav_j^{L}$ randomly} and then downward propagate it through rGBN as in \eqref{DPGDS} to generate {$\thetav_j^{<L}$}.
By assimilating the generated topic weight vectors to the hidden states of the {language model} in each layer, as depicted in \eqref{RNN_hiddenstate}, we generate a corresponding sentence, where we start from a zero hidden state at each layer in the {language model}, and sample words sequentially until the end-of-the-sentence symbol is generated.
The BLEU scores of various methods are shown in Fig. \ref{fig:BLEU}, using the benchmark tool in Texygen \citep{zhu2018texygen}; We show below BLEU-3 and BLEU-4 for {BNC} and defer the analogous plots for IMDB and APNEWS to Appendices \ref{sec:BLEU_IMDB} and \ref{sec:BLEU_APNEWS}, respectively.
Note we set the validation dataset as the ground-truth.
For all datasets, it is clear that rGBN-RNN yields both higher test-BLEU and lower self-BLEU scores than related methods do, indicating the stacked-RNN based language model in rGBN-RNN generalizes well and does not suffer from mode collapse ($i.e.$, low diversity).

\subsection{Qualitative Analysis}

\textbf{Hierarchical structure of language model:} {In Fig. \ref{fig:hidden_states}, we visualize the hierarchical multi-scale structures learned with the language model of rGBN-RNN and that of GBN-RNN, by visualizing the $L_2$-norm of the hidden states in each layer}, {while reading a sentence from the APNEWS validation set as ``\emph{the service employee international union asked why cme group needs tax relief when it is making huge amounts of money?}''
As shown in Fig. \ref{fig:hidden_states} (a), in the bottom hidden layer (h1), the $L_2$ norm sequence varies quickly from word to word, except within short phrases such as ``service employee,'' ``international union,'' and ``tax relief,'' suggesting layer h1 is in charge of capturing short-term local dependencies. By contrast, in the top hidden layer (h3), the $L_2$ norm sequence varies slowly and exhibits semantic/syntactic meaningful long segments, such as ``service employee international union,'' ``asked why cme group needs tax relief,'' ``when it is,'' and ``making huge amounts of,'' suggesting that layer h3 is in charge of capturing long-range dependencies. %
}
Therefore, the language model in rGBN-RNN can allow more specific information to transmit through lower layers, while allowing more general higher level information to transmit through higher layers.
Our proposed model have the ability to learn hierarchical structure of the sequence, despite without designing the multiscale RNNs on purpose like \citet{chung2017hierarchical}.
We also visualize the language model of GBN-RNN in {Fig. \ref{fig:hidden_states} (b)}; with much less smoothly time-evolved deeper layers, GBN-RNN fails to utilize its stacked RNN structure as effectively as rGBN-RNN does.
This suggests that the {language model} is better trained in rGBN-RNN than in GBN-RNN for capturing long-range temporal dependencies, which helps explain why rGBN-RNN exhibits clearly boosted BLEU scores in comparison to GBN-RNN.

\begin{figure*}[!th]
\begin{center}
\includegraphics[%
width=.95\hsize]{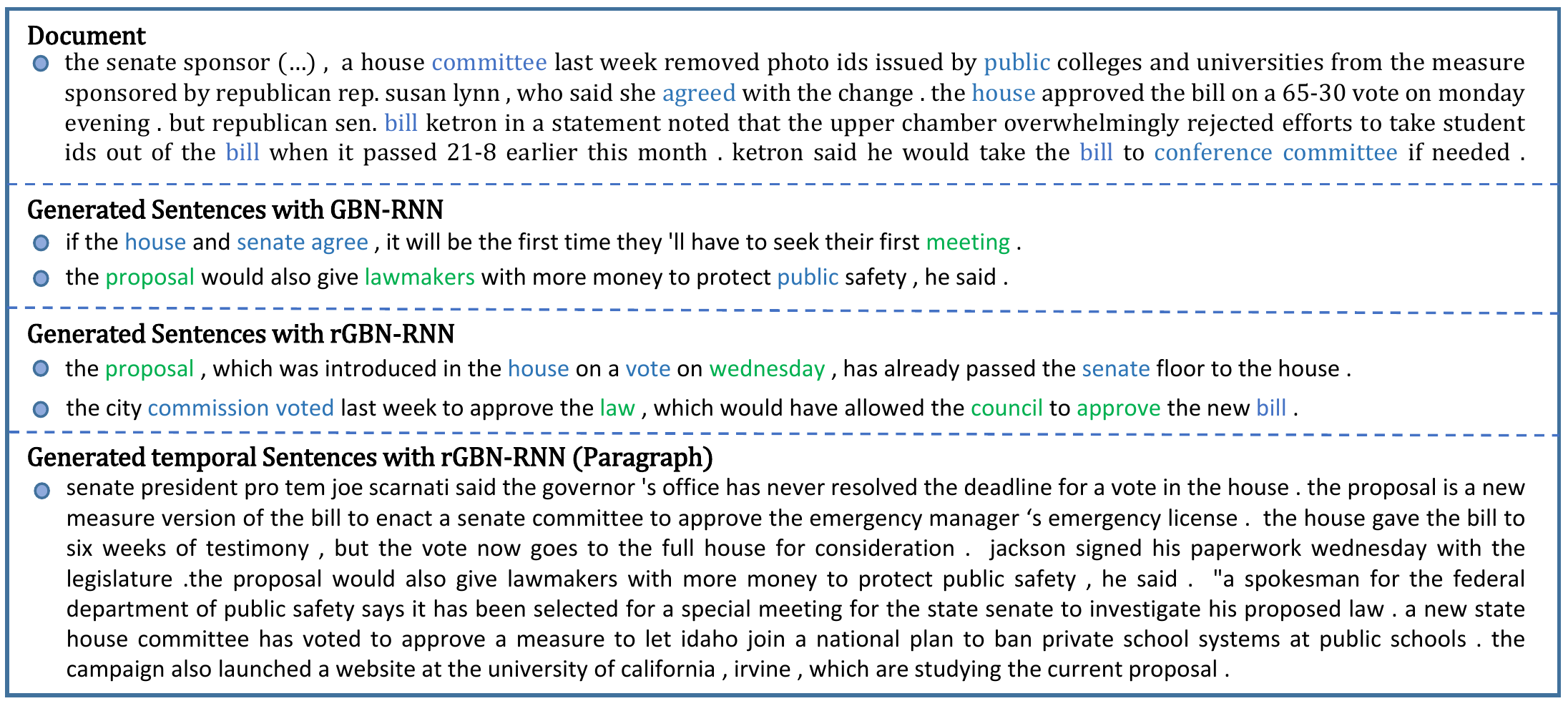}\vspace{-3mm}
\caption{ An example of generated sentences and paragraph conditioned on a document from APNEWS (green denotes novel words, blue the key words in document and generated sentences.) See the Appendix for analogous plots for both IMDB and BNC.
}
\label{fig:sentence_document} \vspace{-2.mm} %
\end{center}
\end{figure*}
\textbf{Sentence generation under topic guidance:} Given the learned rGBN-RNN, we can sample the sentences both conditioning on a single topic of a certain layer and on a combination of the topics from different layers. Shown {in the dotted-line boxes in Fig. \ref{fig:sentence_topic}}, most of the generated sentences conditioned on a single topic or a combination of topics are highly related to the given topics {in terms of their semantical meanings but not necessarily in key words}, indicating the language model is successfully guided by the recurrent hierarchical topics.
These observations suggest that rGBN-RNN has successfully captured syntax and {global} semantics simultaneously for natural language generation. {Similar to Fig. \ref{fig:sentence_topic}, we also provide hierarchical topics and corresponding generated sentences for both IMDB and BNC in Appendix \ref{sec:topics_and_sentences}. Besides, in Appendix \ref{sec:hierarchical_topics}, we provide additional example topic hierarchies and generated sentences  given different topics.}

\textbf{Hierarchical topics:} {We present an example topic hierarchy inferred by a three-layer rGBN-RNN from APNEWS.}
In Fig. \ref{fig:sentence_topic}, we select a large-weighted topic at the top hidden layer and move down the network
to include any lower-layer topics connected to their ancestors with sufficiently large weights.
Horizontal arrows link temporally related topics at the same layer, while top-down arrows link hierarchically related topics across layers.
For example, topic $48$ of layer $3$ on ``budget, lawmakers, gov., revenue'' is related not only in hierarchy to topic $57$ on ``lawmakers, pay, proposal, legislation'' and topic $60$ of the lower layer on ``budget, gov., revenue, vote, costs, mayor,'' but also in time to topic $35$ of the same layer on ``democratic, taxes, proposed, future, state.''
Highly interpretable hierarchical relationships between the topics at different layers, and temporal relationships between the topics at the same layer are captured by rGBN-RNN, and the topics are often quite specific semantically at the bottom layer while becoming increasingly more general when moving upwards.

\textbf{Sentence/paragraph generation conditioning on a paragraph:}
Given the GBN-RNN and rGBN-RNN learned on APNEWS, we further present the generated sentences conditioning on a paragraph, as shown in Fig. \ref{fig:sentence_document}.
We provide analogous plots to Fig. \ref{fig:sentence_document} for both IMDB and BNC in Appendix \ref{sec:paragraphs}.
To randomly generate sentences, we encode %
the paragraph into a hierarchical latent representation and then feed it into the stacked-RNN. Besides, we can generate a paragraph with rGBN-RNN, using its recurrent inference network to encode the paragraph into a dynamic hierarchical latent representation, which is fed into the {language model} to predict the word sequence in each sentence of the input paragraph. It is clear that both the proposed GBN-RNN and rGBN-RNN can successfully capture the key textual information of the input paragraph, and generate diverse realistic sentences.
Interestingly, the proposed rGBN-RNN can generate semantically coherent paragraphs, incorporating contextual information both within and beyond the {sentences}.
Note that with the topics that extract the document-level word concurrence patterns, our proposed models can generate semantically-related words, which may not exist in the original document.

\section{Conclusion}
{We propose a recurrent gamma belief network (rGBN) guided RNN-based language modeling framework, a novel method to jointly learn a neural language model and a deep recurrent topic model. %
For scalable inference, we develop hybrid stochastic gradient MCMC and recurrent autoencoding variational inference, allowing efficient end-to-end training.
Experiments conducted on real world corpora demonstrate that the proposed models outperform a variety of shallow-topic-model-guided RNN-based language models, and effectively generate the sentences from the designated multi-level topics or noise, while inferring interpretable hierarchical latent topic structures of documents and hierarchical multiscale structures of sequences}.
For future work, we plan to extend the proposed models to specific natural language processing tasks, such as machine translation, image paragraph captioning, and text summarization.
{Another promising extension is to replace the stacked-RNN in GBN-RNN or rGBN-RNN with Transformer, %
$i.e.$, constructing a GBN or rGBN guided Transformer as a new larger-context neural language model.}

\section*{Acknowledgements}
B. Chen acknowledges the support of the Program for Young Thousand Talent by Chinese Central Government, the 111 Project (No. B18039), NSFC (61771361), Shaanxi Innovation Team Project, and the Innovation Fund of Xidian University. M. Zhou acknowledges the support of the U.S. National Science Foundation under Grant IIS-1812699.

\bibliography{References}

\begin{thebibliography}{64}
\providecommand{\natexlab}[1]{#1}
\providecommand{\url}[1]{\texttt{#1}}
\expandafter\ifx\csname urlstyle\endcsname\relax
  \providecommand{\doi}[1]{doi: #1}\else
  \providecommand{\doi}{doi: \begingroup \urlstyle{rm}\Url}\fi

\bibitem[Acharya et~al.(2015)Acharya, Ghosh, and Zhou]{GP-DPFA2015}
Acharya, A., Ghosh, J., and Zhou, M.
\newblock Nonparametric {Bayesian} factor analysis for dynamic count matrices.
\newblock In \emph{AISTATS}, 2015.

\bibitem[Ahn et~al.(2017)Ahn, Choi, Parnamaa, and Bengio]{ahn2017a}
Ahn, S., Choi, H., Parnamaa, T., and Bengio, Y.
\newblock A neural knowledge language model.
\newblock \emph{arXiv: Computation and Language}, 2017.

\bibitem[Blei \& Lafferty(2006)Blei and Lafferty]{DTM}
Blei, D.~M. and Lafferty, J.~D.
\newblock Dynamic topic models.
\newblock In \emph{ICML}, 2006.

\bibitem[Blei et~al.(2003)Blei, Ng, and Jordan]{blei2003latent}
Blei, D.~M., Ng, A.~Y., and Jordan, M.~I.
\newblock Latent {D}irichlet allocation.
\newblock \emph{Journal of Machine Learning Research}, 3:\penalty0 993--1022,
  2003.

\bibitem[Cho et~al.(2014)Cho, Merrienboer, Gulcehre, Bougares, and
  Bengio]{Cho2014Learning}
Cho, K., Merrienboer, B.~V., Gulcehre, C., Bougares, F., and Bengio, Y.
\newblock Learning phrase representations using {RNN} encoder-decoder for
  statistical machine translation.
\newblock In \emph{Computer Science}, 2014.

\bibitem[Chung et~al.(2017)Chung, Ahn, and Bengio]{chung2017hierarchical}
Chung, J., Ahn, S., and Bengio, Y.
\newblock Hierarchical multiscale recurrent neural networks.
\newblock In \emph{ICLR}, 2017.

\bibitem[Cong et~al.(2017{\natexlab{a}})Cong, Chen, Liu, and
  Zhou]{cong2017deep}
Cong, Y., Chen, B., Liu, H., and Zhou, M.
\newblock Deep latent {Dirichlet} allocation with topic-layer-adaptive
  stochastic gradient {Riemannian MCMC}.
\newblock In \emph{ICML}, 2017{\natexlab{a}}.

\bibitem[Cong et~al.(2017{\natexlab{b}})Cong, Chen, and Zhou]{cong2017fast}
Cong, Y., Chen, B., and Zhou, M.
\newblock Fast simulation of hyperplane-truncated multivariate normal
  distributions.
\newblock \emph{Bayesian Anal.}, 12\penalty0 (4):\penalty0 1017--1037,
  2017{\natexlab{b}}.

\bibitem[Consortium(2007)]{BNC}
Consortium, B.
\newblock The {B}ritish {N}ational {C}orpus, version 3 ({BNC XML E}dition).
\newblock \url{http://www.natcorp.ox.ac.uk}, 2007.

\bibitem[Dai et~al.(2019)Dai, Yang, Yang, Carbonell, Le, and
  Salakhutdinov]{dai2019transformerxl}
Dai, Z., Yang, Z., Yang, Y., Carbonell, J.~G., Le, Q.~V., and Salakhutdinov, R.
\newblock Transformer-xl: Attentive language models beyond a fixed-length
  context.
\newblock In \emph{ACL}, 2019.

\bibitem[Devlin et~al.(2019)Devlin, Chang, Lee, and Toutanova]{devlin2019bert}
Devlin, J., Chang, M., Lee, K., and Toutanova, K.
\newblock Bert: Pre-training of deep bidirectional transformers for language
  understanding.
\newblock In \emph{north american chapter of the association for computational
  linguistics}, pp.\  4171--4186, 2019.

\bibitem[Dieng et~al.(2017)Dieng, Wang, Gao, and Paisley]{dieng2017topicrnn}
Dieng, A.~B., Wang, C., Gao, J., and Paisley, J.
\newblock Topic{RNN}: A recurrent neural network with long-range semantic
  dependency.
\newblock In \emph{ICLR}, 2017.

\bibitem[Fan et~al.(2020)Fan, Zhang, Wang, and Zhou]{Fan2020Adaptive}
Fan, X., Zhang, Y., Wang, Z., and Zhou, M.
\newblock Adaptive correlated {M}onte {C}arlo for contextual categorical
  sequence generation.
\newblock In \emph{International Conference on Learning Representations}, 2020.

\bibitem[Gan et~al.(2015)Gan, Chen, Henao, Carlson, and Carin]{gan2015scalable}
Gan, Z., Chen, C., Henao, R., Carlson, D., and Carin, L.
\newblock Scalable deep {P}oisson factor analysis for topic modeling.
\newblock In \emph{ICML}, pp.\  1823--1832, 2015.

\bibitem[Gan et~al.(2017)Gan, Gan, He, Pu, Tran, Gao, Carin, and
  Deng]{gan2017semantic}
Gan, Z., Gan, C., He, X., Pu, Y., Tran, K., Gao, J., Carin, L., and Deng, L.
\newblock Semantic compositional networks for visual captioning.
\newblock In \emph{CVPR}, pp.\  1141--1150, 2017.

\bibitem[Gehrmann et~al.(2018)Gehrmann, Deng, and Rush]{gehrmann2018bottom}
Gehrmann, S., Deng, Y., and Rush, A.
\newblock Bottom-up abstractive summarization.
\newblock In \emph{EMNLP}, pp.\  4098--4109, 2018.

\bibitem[Girolami \& Calderhead(2011)Girolami and
  Calderhead]{girolami2011riemann}
Girolami, M.~A. and Calderhead, B.
\newblock Riemann manifold {Langevin and Hamiltonian Monte Carlo} methods.
\newblock \emph{Journal of The Royal Statistical Society Series B-statistical
  Methodology}, 73\penalty0 (2):\penalty0 123--214, 2011.

\bibitem[Graves(2013)]{graves2013generating}
Graves, A.
\newblock Generating sequences with recurrent neural networks.
\newblock \emph{arXiv: Neural and Evolutionary Computing}, 2013.

\bibitem[{Graves} et~al.(2013){Graves}, {Mohamed}, and {Hinton}]{alex2013}
{Graves}, A., {Mohamed}, A., and {Hinton}, G.
\newblock Speech recognition with deep recurrent neural networks.
\newblock In \emph{ICASSP}, pp.\  6645--6649, 2013.

\bibitem[Griffiths \& Steyvers(2004)Griffiths and
  Steyvers]{griffiths2004finding}
Griffiths, T.~L. and Steyvers, M.
\newblock Finding scientific topics.
\newblock \emph{Proceedings of the National Academy of Sciences}, 101:\penalty0
  5228--5235, 2004.

\bibitem[Griffiths et~al.(2004)Griffiths, Steyvers, Blei, and
  Tenenbaum]{griffiths2004integrating}
Griffiths, T.~L., Steyvers, M., Blei, D.~M., and Tenenbaum, J.~B.
\newblock Integrating topics and syntax.
\newblock In \emph{NeurIPS}, pp.\  537--544, 2004.

\bibitem[Guo et~al.(2018)Guo, Chen, Zhang, and Zhou]{guo2018deep}
Guo, D., Chen, B., Zhang, H., and Zhou, M.
\newblock Deep {P}oisson gamma dynamical systems.
\newblock In \emph{NeurIPS}, pp.\  8451--8461, 2018.

\bibitem[Hochreiter \& Schmidhuber(1997)Hochreiter and
  Schmidhuber]{hochreiter1997long}
Hochreiter, S. and Schmidhuber, J.
\newblock Long short-term memory.
\newblock \emph{Neural Computation}, 9\penalty0 (8):\penalty0 1735--1780, 1997.

\bibitem[Hoffman et~al.(2013)Hoffman, Blei, Wang, and
  Paisley]{hoffman2013stochastic}
Hoffman, M.~D., Blei, D.~M., Wang, C., and Paisley, J.
\newblock Stochastic variational inference.
\newblock \emph{The Journal of Machine Learning Research}, 14\penalty0
  (1):\penalty0 1303--1347, 2013.

\bibitem[Kingma \& Ba(2015)Kingma and Ba]{kingma2015adam}
Kingma, D.~P. and Ba, J.
\newblock Adam: A method for stochastic optimization.
\newblock In \emph{ICLR}, 2015.

\bibitem[Kingma \& Welling(2013)Kingma and Welling]{kingma2013auto}
Kingma, D.~P. and Welling, M.
\newblock Auto-encoding variational bayes.
\newblock In \emph{ICLR}, 2013.

\bibitem[Klein \& Manning(2003)Klein and Manning]{Manning2003Accurate}
Klein, D. and Manning, C.~D.
\newblock Accurate unlexicalized parsing.
\newblock In \emph{Meeting of the Association for Computational Linguistics},
  2003.

\bibitem[Lau et~al.(2017)Lau, Baldwin, and Cohn]{lau2017topically}
Lau, J.~H., Baldwin, T., and Cohn, T.
\newblock Topically driven neural language model.
\newblock In \emph{meeting of the association for computational linguistics},
  pp.\  355--365, 2017.

\bibitem[Li et~al.(2015)Li, Chen, Carlson, and Carin]{li2015preconditioned}
Li, C., Chen, C., Carlson, D., and Carin, L.
\newblock Preconditioned stochastic gradient {L}angevin dynamics for deep
  neural networks.
\newblock \emph{arXiv}, 2015.

\bibitem[Ma et~al.(2015)Ma, Chen, and Fox]{ma2015complete}
Ma, Y., Chen, T., and Fox, E.
\newblock A complete recipe for stochastic gradient {MCMC}.
\newblock In \emph{NIPS}, pp.\  2899--2907, 2015.

\bibitem[Maas et~al.(2011)Maas, Daly, Pham, Dan, Ng, and
  Potts]{Maas2011Learning}
Maas, A.~L., Daly, R.~E., Pham, P.~T., Dan, H., Ng, A.~Y., and Potts, C.
\newblock Learning {W}ord {V}ectors for {S}entiment {A}nalysis.
\newblock In \emph{Meeting of the Association for Computational Linguistics:
  Human Language Technologies}, 2011.

\bibitem[Mao et~al.(2015)Mao, Xu, Yang, Wang, Huang, and Yuille]{mao2015deep}
Mao, J., Xu, W., Yang, Y., Wang, J., Huang, Z., and Yuille, A.~L.
\newblock Deep captioning with multimodal recurrent neural networks m-{RNN}.
\newblock In \emph{ICLR}, 2015.

\bibitem[Miao et~al.(2017)Miao, Grefenstette, and Blunsom]{miao2017discovering}
Miao, Y., Grefenstette, E., and Blunsom, P.
\newblock Discovering discrete latent topics with neural variational inference.
\newblock In \emph{ICML}, pp.\  2410--2419, 2017.

\bibitem[Mikolov \& Zweig(2012)Mikolov and Zweig]{mikolov2012context}
Mikolov, T. and Zweig, G.
\newblock Context dependent recurrent neural network language model.
\newblock In \emph{SLT}, pp.\  234--239, 2012.

\bibitem[Mikolov et~al.(2010)Mikolov, Karafiat, Burget, Cernocky, and
  Khudanpur]{mikolov2010recurrentNEW}
Mikolov, T., Karafiat, M., Burget, L., Cernocky, J., and Khudanpur, S.
\newblock Recurrent neural network based language model.
\newblock In \emph{Interspeech}, 2010.

\bibitem[Mikolov et~al.(2011)Mikolov, Kombrink, Burget, Cernocky, and
  Khudanpur]{mikolov2011extensions}
Mikolov, T., Kombrink, S., Burget, L., Cernocky, J., and Khudanpur, S.
\newblock Extensions of recurrent neural network language model.
\newblock In \emph{ICASSP}, pp.\  5528--5531, 2011.

\bibitem[Mnih \& Gregor(2014)Mnih and Gregor]{mnih2014neural}
Mnih, A. and Gregor, K.
\newblock Neural variational inference and learning in belief networks.
\newblock In \emph{ICML}, pp.\  1791--1799, 2014.

\bibitem[Patterson \& Teh(2013)Patterson and Teh]{patterson2013stochastic}
Patterson, S. and Teh, Y.~W.
\newblock Stochastic gradient {R}iemannian {L}angevin dynamics on the
  probability simplex.
\newblock In \emph{NIPS}, pp.\  3102--3110, 2013.

\bibitem[Radford et~al.(2018)Radford, Narasimhan, Salimans, and
  Sutskever]{radford2018improving}
Radford, A., Narasimhan, K., Salimans, T., and Sutskever, I.
\newblock Improving language understanding by generative pre-training.
\newblock 2018.

\bibitem[Radford et~al.(2019)Radford, Wu, Child, Luan, Amodei, and
  Sutskever]{radford2019language}
Radford, A., Wu, J., Child, R., Luan, D., Amodei, D., and Sutskever, I.
\newblock Language models are unsupervised multitask learners.
\newblock 2019.

\bibitem[Rennie et~al.(2017)Rennie, Marcheret, Mroueh, Ross, and
  Goel]{rennie2017self}
Rennie, S.~J., Marcheret, E., Mroueh, Y., Ross, J., and Goel, V.
\newblock Self-critical sequence training for image captioning.
\newblock In \emph{Proceedings of the IEEE Conference on Computer Vision and
  Pattern Recognition}, pp.\  7008--7024, 2017.

\bibitem[Rezende et~al.(2014)Rezende, Mohamed, and
  Wierstra]{rezende2014stochastic}
Rezende, D.~J., Mohamed, S., and Wierstra, D.
\newblock Stochastic backpropagation and approximate inference in deep
  generative models.
\newblock In \emph{ICML}, pp.\  1278--1286, 2014.

\bibitem[Rush et~al.(2015)Rush, Chopra, and Weston]{rush2015a}
Rush, A.~M., Chopra, S., and Weston, J.
\newblock A neural attention model for abstractive sentence summarization.
\newblock In \emph{EMNLP}, pp.\  379--389, 2015.

\bibitem[Schein et~al.(2016)Schein, Wallach, and
  Zhou]{ScheinWallachZhou_PGDS_2016}
Schein, A., Wallach, H., and Zhou, M.
\newblock Poisson--gamma dynamical systems.
\newblock In \emph{Neural Information Processing Systems}, 2016.

\bibitem[Srivastava \& Sutton(2017)Srivastava and
  Sutton]{srivastava2017autoencoding}
Srivastava, A. and Sutton, C.
\newblock Autoencoding variational inference for topic models.
\newblock In \emph{ICLR}, 2017.

\bibitem[Srivastava et~al.(2013)Srivastava, Salakhutdinov, and
  Hinton]{srivastava2013modeling}
Srivastava, N., Salakhutdinov, R., and Hinton, G.~E.
\newblock Modeling documents with deep {B}oltzmann machines.
\newblock In \emph{Uncertainty in Artificial Intelligence}, 2013.

\bibitem[Sutskever et~al.(2014)Sutskever, Vinyals, and
  Le]{sutskever2014sequence}
Sutskever, I., Vinyals, O., and Le, Q.~V.
\newblock Sequence to sequence learning with neural networks.
\newblock In \emph{Advances in neural information processing systems}, pp.\
  3104--3112, 2014.

\bibitem[Teh et~al.(2006)Teh, Jordan, Beal, and Blei]{Teh2006Hierarchical}
Teh, Y.~W., Jordan, M.~I., Beal, M.~J., and Blei, D.~M.
\newblock Hierarchical {D}irichlet processes.
\newblock \emph{Publications of the American Statistical Association},
  101\penalty0 (476):\penalty0 1566--1581, 2006.

\bibitem[Tian \& Cho(2016)Tian and Cho]{Tian2016Larger}
Tian, W. and Cho, K.
\newblock Larger-context language modelling with recurrent neural network.
\newblock In \emph{Meeting of the Association for Computational Linguistics},
  2016.

\bibitem[Vaswani et~al.(2017)Vaswani, Shazeer, Parmar, Uszkoreit, Jones, Gomez,
  Kaiser, and Polosukhin]{vaswani2017attention}
Vaswani, A., Shazeer, N., Parmar, N., Uszkoreit, J., Jones, L., Gomez, A.~N.,
  Kaiser, L., and Polosukhin, I.
\newblock Attention is all you need.
\newblock In \emph{Neural Information Processing Systems}, pp.\  6000--6010,
  2017.

\bibitem[Vinyals et~al.(2015)Vinyals, Toshev, Bengio, and
  Erhan]{vinyals2015show}
Vinyals, O., Toshev, A., Bengio, S., and Erhan, D.
\newblock Show and tell: A neural image caption generator.
\newblock In \emph{CVPR}, 2015.

\bibitem[Wallach(2006)]{wallach2006topic}
Wallach, H.~M.
\newblock Topic modeling: beyond bag-of-words.
\newblock In \emph{ICML}, pp.\  977--984, 2006.

\bibitem[Wang et~al.(2019{\natexlab{a}})Wang, Chen, Xiao, and
  Zhou]{wang2019convolutional}
Wang, C., Chen, B., Xiao, S., and Zhou, M.
\newblock Convolutional {P}oisson gamma belief network.
\newblock In \emph{International Conference on Machine Learning}, pp.\
  6515--6525, 2019{\natexlab{a}}.

\bibitem[Wang et~al.(2018)Wang, Gan, Wang, Shen, Huang, Ping, Satheesh, and
  Carin]{wang2018topic}
Wang, W., Gan, Z., Wang, W., Shen, D., Huang, J., Ping, W., Satheesh, S., and
  Carin, L.
\newblock Topic compositional neural language model.
\newblock In \emph{AISTATS}, pp.\  356--365, 2018.

\bibitem[Wang et~al.(2019{\natexlab{b}})Wang, Gan, Xu, Zhang, Wang, Shen, Chen,
  and Carin]{wang2019topic_guided}
Wang, W., Gan, Z., Xu, H., Zhang, R., Wang, G., Shen, D., Chen, C., and Carin,
  L.
\newblock Topic-guided variational autoencoders for text generation.
\newblock In \emph{NAACL}, 2019{\natexlab{b}}.

\bibitem[Welling \& Teh(2011)Welling and Teh]{welling2011bayesian}
Welling, M. and Teh, Y.~W.
\newblock {B}ayesian learning via stochastic gradient {L}angevin dynamics.
\newblock In \emph{ICML}, pp.\  681--688, 2011.

\bibitem[Xu et~al.(2015)Xu, Ba, Kiros, Cho, Courville, Salakhutdinov, Zemel,
  and Bengio]{xu2015show}
Xu, K., Ba, J., Kiros, R., Cho, K., Courville, A.~C., Salakhutdinov, R., Zemel,
  R.~S., and Bengio, Y.
\newblock Show, attend and tell: Neural image caption generation with visual
  attention.
\newblock In \emph{ICML}, 2015.

\bibitem[Zhang et~al.(2018)Zhang, Chen, Guo, and Zhou]{Zhang2018WHAI}
Zhang, H., Chen, B., Guo, D., and Zhou, M.
\newblock {WHAI: W}eibull hybrid autoencoding inference for deep topic
  modeling.
\newblock In \emph{ICLR}, 2018.

\bibitem[Zhao et~al.(2018)Zhao, Du, Buntine, and Zhou]{zhao2018dirichlet}
Zhao, H., Du, L., Buntine, W., and Zhou, M.
\newblock Dirichlet belief networks for topic structure learning.
\newblock In \emph{Neural Information Processing Systems}, pp.\  7955--7966,
  2018.

\bibitem[Zhou \& Carin(2015)Zhou and Carin]{NBP2012}
Zhou, M. and Carin, L.
\newblock Negative binomial process count and mixture modeling.
\newblock \emph{IEEE Trans. Pattern Analysis and Machine Intelligence},
  37\penalty0 (2):\penalty0 307--320, 2015.

\bibitem[Zhou et~al.(2012)Zhou, Hannah, Dunson, and Carin]{zhou2012beta}
Zhou, M., Hannah, L., Dunson, D.~B., and Carin, L.
\newblock Beta-negative binomial process and {P}oisson factor analysis.
\newblock In \emph{AISTATS}, pp.\  1462--1471, 2012.

\bibitem[Zhou et~al.(2015)Zhou, Cong, and Chen]{zhou2015poisson}
Zhou, M., Cong, Y., and Chen, B.
\newblock The {P}oisson gamma belief network.
\newblock In \emph{NIPS}, pp.\  3025--3033, 2015.

\bibitem[Zhou et~al.(2016)Zhou, Cong, and Chen]{GBN}
Zhou, M., Cong, Y., and Chen, B.
\newblock Augmentable gamma belief networks.
\newblock \emph{J. Mach. Learn. Res.}, 17\penalty0 (163):\penalty0 1--44, 2016.

\bibitem[Zhu et~al.(2018)Zhu, Lu, Zheng, Guo, Zhang, Wang, and
  Yu]{zhu2018texygen}
Zhu, Y., Lu, S., Zheng, L., Guo, J., Zhang, W., Wang, J., and Yu, Y.
\newblock Texygen: A benchmarking platform for text generation models.
\newblock \emph{SIGIR}, 2018.

\end{thebibliography}
\bibliographystyle{icml2020}
\clearpage
\appendix

\onecolumn
\section{GBN-RNN}\label{sec:GBN-RNN}

\textbf{GBN-RNN:} $\{y_{1:T},{\dv}\}$ denotes a sentence-context pair,
where ${\dv}\in \mathbb{Z}_+^{V_c}$ represents the document-level context as a word frequency count vector, the $v$th element of which counts the number of times the $v$th word in the vocabulary appears in the document excluding sentence $y_{1:T}$.
The hierarchical model of an $L$-hidden-layer GBN, from top to bottom, is expressed as
\begin{align}\label{WHAI_decoder}
 &\thetav^{L} \sim \mbox{Gam}\left(\rrv,c^{L+1} \right),\, \ldots,\thetav^{l} \sim \mbox{Gam}\left(\Phimat^{l+1} \thetav^{l+1} ,c^{l+1} \right),\, \ldots, \nonumber \\
 & \thetav^{1} \!\sim \mbox{Gam}\left(\Phimat^{2} \thetav^{2} ,c^{2} \right),~ \dv \!\sim \!\mbox{Pois} \left(\Phimat^{1} \thetav^{1} \right).
\end{align}
 The stacked-RNN based language model described in \eqref{RNN_hiddenstate} is also used in GBN-RNN.

\textbf{Statistical inference:}
 To infer GBN-RNN, we consider a hybrid of stochastic gradient MCMC \citep{welling2011bayesian,patterson2013stochastic,li2015preconditioned,ma2015complete,cong2017deep}, used for the GBN topics $\phiv_k^{l}$, and auto-encoding variational inference \citep{kingma2013auto,rezende2014stochastic}, used for the parameters of both the inference network (encoder) and RNN. More specifically,
GBN-RNN generalizes Weibull hybrid auto-encoding inference (WHAI) of \citet{Zhang2018WHAI}: it uses a deterministic-downward-stochastic-upward inference network to encode the bag-of-words representation of $ \dv$ into the latent topic-weight variables $\thetav^{l}$ across all hidden layers, which are fed into not only GBN to reconstruct $ \dv$, but also a stacked RNN in language model, as shown in \eqref{RNN_hiddenstate}, to predict the word sequence in $y_{1:T}$.
The topics $\phiv_k^{l}$ can be sampled with topic-layer-adaptive stochastic gradient Riemannian (TLASGR) MCMC, whose details can be found in \citet{cong2017deep,Zhang2018WHAI}, omitted here for brevity.
Given the sampled topics $\phiv_k^{l}$,
the joint marginal likelihood of $\{y_{1:T}, \dv\}$ is defined as
\begin{equation}\label{joint_like}
p\left(y_{1:T}, \dv\given {\{\Phimat^{l}\}_l} \right) \! = \! \displaystyle{\int p\left( \dv \given \Phimat^{1}\thetav^{1}\right) \left[ \prod\limits_{t = 1}^{T} \! p\left( {y_t}\given y_{1:t-1},\thetav^{1:L} \right)\right]\left[\prod\limits_{l = 1}^{L} \! {p\left(\thetav^{l}\given \Phimat^{l + 1}\thetav^{l+1} \right)}\right] d_{\thetav^{1:L}} }.
\end{equation}
For efficient inference, an inference network as $Q=\prod_{l=1}^L q(\thetav^{l}\given \dv, \Phimat^{l+1}\thetav^{l+1})$ is used to provide an ELBO of the log joint marginal likelihood as
 \begin{equation}\label{ELBO-of-our model1}
L(y_{1:T}, \dv) = \mathbb{E}_Q\left[ \ln p\left( \dv \given \, \Phimat^{1}\thetav^{1} \right)+ \sum \limits_{t=1}^T \ln p\left( {y_t}\given y_{1:t-1},\thetav^{1:L}\right)\right] - \sum \limits_{l=1}^L \mathbb{E}_Q \left[ \ln \frac{ q\left( \thetav^{l} \given \dv, \Phimat^{l+1}\thetav^{l+1}\right)} {p \left( \thetav^{l} \,\given \, \Phimat^{l+1}\thetav^{l+1} \right)} \right]
\end{equation}
and the training is performed by maximizing $\mathbb{E}_{p_{\text{data}}(y_{1:T}, \dv)}[L(y_{1:T}, \dv)]$;
following { \citet{Zhang2018WHAI}, we define
 $q(\thetav^{l} \given \dv, \Phimat^{l+1} ,\thetav^{l+1}) = \mbox{Weibull}(\kv^{l}+\Phimat^{l+1} \thetav^{l+1},\lambdav^{l})$, where both $\kv^{l}$ and $\lambdav^{l}$ are deterministically transformed from~$ \dv$ using neural networks.
Distinct from a usual variational auto-encoder whose inference network has a pure bottom-up structure, the inference network here has a determistic-upward--stoachstic-downward ladder structure \citep{Zhang2018WHAI}.

\section{TLASGR-MCMC for rGBN-RNN}\label{sec:SGMCMC for GBN-RNN}
To allow for scalable inference, we apply the TLASGR-MCMC algorithm \citep{cong2017deep, Zhang2018WHAI,guo2018deep}, which can be used to sample simplex-constrained global parameters \cite{cong2017fast} in a mini-batch based manner. It improves its sampling efficiency via the
 use of the Fisher information matrix (FIM) \cite{girolami2011riemann}, with
adaptive step-sizes for the latent factors and transition matrices of different layers.
In this section, we discuss how to update the global parameters $\{\Phimat^{l},\Pimat^{l}\}_{l=1}^L$ of rGBN in detail and give a complete one in Algorithm in \ref{Algorithm}.

{{\bf{Sample the auxiliary counts:}}
This step is about the ``backward'' and ``upward'' pass. Let us denote $Z_{\cdotv kj}^{{l}} = \sum_{{k_l} = 1}^{{K_l}} {Z_{{k_l}kj}^{ {l}}} $, ~$Z_{\cdotv {k},{J+1}}^{{l}}=0$, and $x_{kj}^{(1,1)}=d_{vj}$, where $\dv_{j}=\{d_{1j},..,d_{vj},..,d_{{V_c}j}\}$ is the same as in %
\eqref{DPGDS}. Working backward for $j = J, . . . , 1$ and upward for $l = 1,...,L$, we draw
\begin{align}\label{Multi_Phi_Theta}
 & ( {A_{k1j}^{l},...,A_{k{K_l}j}^{l}} )\sim \mbox{Multi}\left(x_{kj}^{(l,l)};\frac{{\phi _{k{1}}^{l}\theta _{{1}j}^{l}}}{{\sum\nolimits_{{k_l} = 1}^{{K_l}} {\phi_{k{k_l}}^{l}\theta _{{k_l}j}^{l}} }},...,\frac{{\phi _{k{K_l}}^{l}\theta _{{K_l}j}^{l}}}{{\sum\nolimits_{{k_l} = 1}^{{K_l}} {\phi _{k{k_l}}^{l}\theta_{{k_l}j}^{l}} }}\right),\\
\label{auxiliary_variables}
 & x_{kj}^{{l+1 }}\sim \textrm{CRT}\left[ {A_{\cdotv kj}^{l}+Z_{\cdotv k,j+1}^{{l}},{\tau_0}\left( {\sum\nolimits_{{k_{l + 1}} = 1}^{{K_{l + 1}}} {\phi _{k{k_{l + 1}}}^{ {l + 1}}\theta _{{k_{l + 1}}j}^{l + 1}} + \sum\nolimits_{{k_l} = 1}^{{K_l}} {\pi _{k{k_1}}^{l}\theta _{{k_1},j - 1}^{l }} } \right)} \right].
\end{align}
Note that via the deep structure, the latent counts $x_{kj}^{l+1}$ will be influenced by the effects from both time $j-1$ at layer~$l$ and time $j$ at layer $l+1$.
With %
$p_1 := \sum\nolimits_{{k_l} = 1}^{{K_l}} {\pi_{k{k_l}}^{l}\theta_{{k_l}j - 1}^{l}}$ and
$p_2 := \sum\nolimits_{{k_{l+1}} = 1}^{{K_{l+1}}} {\phi_{k{k_{l+1}}}^{l+1}\theta_{{k_{l+1}}j}^{l+1}}
$, we can sample the latent counts at layer~$l$ and $l+1$ by
\begin{equation}\label{Two_Poisson}
 (x_{kj}^{{l+1,l}},x_{kj}^{{l+1,l+1}})\sim \textrm{Multi}\left(x_{kj}^{{l+1}},{p_1}/{(p_1+p_2)},{p_2}/{(p_1+p_2)}\right),
\end{equation}
and then draw
\begin{equation}\label{Multi_Pi_Theta}
( {Z_{k1j}^{{l}},...,Z_{k{K_l}j}^{{l}}} )\sim \textrm{Multi} \left( {x_{kj}^{{l+1,l}};\frac{{\pi_{k1}^{l}\theta _{1,j - 1}^{l }}}{{\sum\nolimits_{{k_l} = 1}^{{K_l}} {\pi _{{k}{k_l}}^{l}\theta _{k_l,j - 1}^{l}} }},...,\frac{{\pi _{k{K_l}}^{l}\theta _{{K_l},j - 1}^{l}}}{{\sum\nolimits_{{k_l} = 1}^{{K_l}} {\pi _{k{k_l}}^{l}\theta _{{k_l},j - 1}^{( l )}} }}} \right).
\end{equation}}

In rGBN, the prior and the likelihood of $\{\Phimat^{l}\}_{l=1}^L$ is very similar to $\{\Pimat^{l}\}_{l=1}^L$, so we also apply the TLASGR-MCMC sampling algorithm on both of them conditioned on the auxiliary counts.

{\bf{Sample the hierarchical components $\{\Phimat^{l}\}_{l=1}^L$: }}For $\phiv_k^{l}$, the $k$th column of the loading matrix $\Phimat^{l}$ of layer $l$, its sampling can be efficiently realized as
\begin{align}\label{TLASGR update_Phi}
\left( {\phiv_k^{l}} \right)_{n + 1} \! = & \! \bigg[ \! \left( {\phiv_k^{l}} \right)_n \! + \! \frac{\varepsilon _n}{P_k^{l}} \! \left[ \left(\rho \tilde \Av_{:k\cdotv}^{l} \! + \! \eta_{0}^{l}\right) \! - \! \left(\rho \tilde A_{\cdotv k\cdotv}^{l} \! + \! K_{l-1}\eta_{0}^{l} \right) \! \left( {\phiv_k^{l}} \right)_n \right] \nonumber \\
& + \mathcal{N} \left( 0, \frac{2 \varepsilon _n}{P_k^{l}}\left[ \mbox{diag}(\phiv_k^{l})_n - (\phiv_k^{l})_n (\phiv_k^{l})_n^T \right] \right) \bigg]_\angle,
\end{align}
where $P_k^{l}$ is calculated using the estimated FIM, $\tilde A_{{k_l}j\cdotv}^{{l}} = \sum_{{j} = 1}^{J} {A_{{k_l}kj}^{ {l}}}, {\tilde \Av_{:k\cdotv}^{l }} = \{\tilde A_{{1}j\cdotv}^{{l}},\cdots, \tilde A_{{K_l}j\cdotv}^{{l}} \}^{T} $ and ${\tilde A_{\cdotv k\cdotv}^{l}}= \sum_{{k_l} = 1}^{k_l} \tilde A_{{k_l}j\cdotv}^{{l}} $, ${A_{{k_l}kj}^{ {l}}}$ comes from the augmented latent
counts $A^{l}$ in \eqref{Multi_Phi_Theta},
$\eta_{0}^{l}$ denote the prior of ${\phiv_k^{l}}$,
and $[\cdot]_\angle$ denotes the simplex constraint.

{\bf{Sample the transmission matrix $\{\Pimat^{l}\}_{l=1}^L$:}}
For $\piv_k^{l}$, the $k$th column of the transition matrix $\Pimat^{l}$ of layer $l$, %
 its sampling can be efficiently realized as
\begin{align}\label{TLASGR Pi}
\left( {\piv_k^{l}} \right)_{n + 1} \! = & \! \bigg[ \! \left( {\piv_k^{l}} \right)_n \! + \! \frac{\varepsilon _n}{M_k^{l}} \! \left[ \left(\rho \tilde \Zv_{:k\cdotv}^{l} \! + \! \etav_{:k}^{l}\right) \! - \! \left(\rho \tilde Z_{\cdotv k\cdotv}^{l} \! + \! \eta_{.k}^{l} \right) \! \left( {\piv_k^{l}} \right)_n \right] \nonumber \\
& + \mathcal{N} \left( 0, \frac{2 \varepsilon _n}{M_k^{l}}\left[ \mbox{diag}(\piv_k^{l})_n - (\piv_k^{l})_n (\piv_k^{l})_n^T \right] \right) \bigg]_\angle,
\end{align}
where $M_k^{l}$ is calculated using the estimated FIM, $\tilde Z_{{k_l}j\cdotv}^{{l}} = \sum_{{j} = 1}^{J} {Z_{{k_l}kj}^{ {l}}}, {\tilde \Zv_{:k\cdotv}^{l }} = \{\tilde Z_{{1}j\cdotv}^{{l}},\cdots, \tilde Z_{{K_l}j\cdotv}^{{l}} \}^{T} $ and ${\tilde Z_{\cdotv k\cdotv}^{l}}= \sum_{{k_l} = 1}^{k_l} \tilde Z_{{k_l}j\cdotv}^{{l}} $, ${Z_{{k_l}kj}^{ {l}}}$ comes from the augmented latent
counts $Z^{l}$ in \eqref{Multi_Pi_Theta}, and ${\left[ . \right]_\angle }$ denotes a simplex constraint, and ${\etav_{:k}^{l}}$ denotes the prior of ${\piv_k^{l}}$, more details about TLASGR-MCMC for our proposed model can be found in %
\citet{cong2017deep}.

\section{Datasets}\label{sec:data}
 We consider three publicly available corpora\footnote{\url{https://ibm.ent.box.com/s/ls61p8ovc1y87w45oa02zink2zl7l6z4}}. APNEWS is a collection of Associated Press %
 news articles from 2009 to 2016, IMDB is a set of movie reviews collected by \citet{Maas2011Learning}, and BNC is the written portion of the British National Corpus \citep{BNC}. %
Following the preprocessing steps in \citet{lau2017topically}, we tokenize words and sentences using Stanford CoreNLP \citep{Manning2003Accurate}, lowercase all word tokens, and filter out word tokens that occur less than 10 times.
For the topic model, we additionally exclude stopwords\footnote{We use Mallet's stopword list: \url{https://github.com/mimno/Mallet/tree/master/stoplists}} and the top $0.1\%$ most frequent words. All these corpora are partitioned into training, validation, and testing sets, whose summary statistics are provided in Table \ref{Tab:Summary of datasets} of the Appendix.

\begin{table*}[h]
\linespread{1.25}
\footnotesize
\centering
 \caption{ Summary statistics for the datasets.}%
\begin{tabular}{c|cc|ccc|ccc|ccc}
\toprule[1pt]
\multirow{2}{*}{{\textbf{Dataset}}} &\multicolumn{2}{c|}{{\textbf{Vocubalry}}} & \multicolumn{3}{c|}{{\textbf{Training}}} & \multicolumn{3}{c|}{{\textbf{Validation}}}& \multicolumn{3}{c}{{\textbf{Testing}}}\\
\cline{2-3}\cline{4-6} \cline{7-9}\cline{10-12}
 & LM& TM& Docs & Sents & Tokens & Docs & Sents & Tokens& Docs & Sents & Tokens \\
\hline
{APNEWS} & 34231 & 32169 &50K &0.8M & 15M & 2K & 33K & 0.6M & 2K & 32K & 0.6M \\
{IMDB} & 36009 & 34925 &75K & 1.1M& 20M & 12.5K & 0.18M & 0.3M & 12.5K & 0.18M & 0.3M \\
{BNC} & 43703& 41552 &15K & 1M & 18M & 1K & 57K & 1M & 1K & 66K & 1M\\
\bottomrule
 \end{tabular}\label{Tab:Summary of datasets}	
\vspace{-2mm}
\end{table*}

{\section{Complexity of rGBN-RNN}\label{sec:parameters}
The proposed rGBN-RNN consists of both language model and topic model components. For the topic model component, there are
the global parameters of rGBN (decoder), including $\{ \Phimat^{l},\Pimat^{l} \}_{l=1}^{L}$ in \eqref{DPGDS} , and the parameters of the recurrent variational inference network (encoder), consisting of $\mathrm{RNN}_{\mathrm{sent}}^{l}$, $f_{\kv}^{l}$, and $f_{\lambdav}^{l}$ in \eqref{update_network_parameters}. The language model component is parameterized by $\mathrm{LSTM}_{\mathrm{word}}^{l}$ in \eqref{RNN_hiddenstate} and the coupling vectors $g^{l}$ described in \eqref{coupling_vector}.
We summarize in Table \ref{Tab:complexity} the complexity of rGBN-RNN (ignoring all bias terms), where $V$ denotes the vocabulary size of the language model, $E$ the dimension of word embedding vectors, $V_c$ the size of the vocabulary of the topic model that excludes stop words, $H_l^{w}$ the number of hidden units of the word-level LSTM at layer $l$ (stacked-RNN language model), $H_l^{s}$ the number of hidden units of the sentence-level RNN at layer $l$ ( recurrent variational inference network), and $K_l$ the number of topics at layer $l$.}
Comparison of the number of parameters between various %
language models is provided in Table \ref{Tab:perplexity and topic coherence}. %
\begin{table*}[t]\huge
\centering
\caption{ Complexity of the three-layer rGBN-RNN.}
\resizebox{1\textwidth}{!}{
\begin{tabular}{c|c|c|c|c|c|c|c}%
\toprule[1pt]
\textbf{Component} & \multicolumn{2}{c|}{Language Model } & \multicolumn{5}{c}{Topic Model} \\ \cline{1-3} \cline{4-8}
\textbf{Param} &$\mathrm{LSTM}_{\mathrm{word}}^{l}$ in \eqref{RNN_hiddenstate} & $g^{l} $ in \eqref{coupling_vector} & $\Phimat^{l}$ in \eqref{DPGDS} &$ \Pimat^{l}$ in \eqref{DPGDS} &$\mathrm{RNN}_{\mathrm{sent}}^{l}$ in \eqref{update_network_parameters} &$f_{\kv}^{l}$ in \eqref{update_network_parameters} &$f_{\lambdav}^{l}$ in \eqref{update_network_parameters}\\
\hline
Layer1 &$O(4\times (E+H_1^{w})\times H_1^{w})$ &$O(3\times(K_1+H_1^{w})\times H_1^{w})$ &$O(V_c \times K_1)$ &$O(K_1 \times K_1)$ & $O((V_c+H_1^{s})\times H_1^{s})$ & $ O(H_1^{s})$ & $O(K_1 \times H_1^{s})$ \\
Layer2 &$O(4\times (H_1^{w}+H_2^{w})\times H_2^{w})$ &$O(3\times(K_2+H_2^{w})\times H_2^{w})$ &$O(K_1 \times K_2)$ &$O(K_2 \times K_2)$ & $O((H_1^{s}+H_2^{s})\times H_2^{s})$ & $O(H_2^{s})$ & $O(K_2 \times H_2^{s} )$\\
Layer3 & $O(4\times (H_2^{w}+H_3^{w})\times H_3^{w})$ &$O(3\times(K_3+H_3^{w})\times H_3^{w})$ &$O(K_2 \times K_3)$ &$O(K_3 \times K_3)$ &$O((H_2^{s}+H_3^{s})\times H_3^{s})$ & $O(H_3^{s})$ &$O(K_3 \times H_3^{s} )$\\
\bottomrule
\end{tabular}\label{Tab:complexity}
}
\label{tab2}
\end{table*}
}

\section{BLEU scores for IMDB}\label{sec:BLEU_IMDB}\vspace{-2mm}
\begin{figure*}[ht!]
\begin{center}
\includegraphics[height=4.5cm,width=6.6cm]{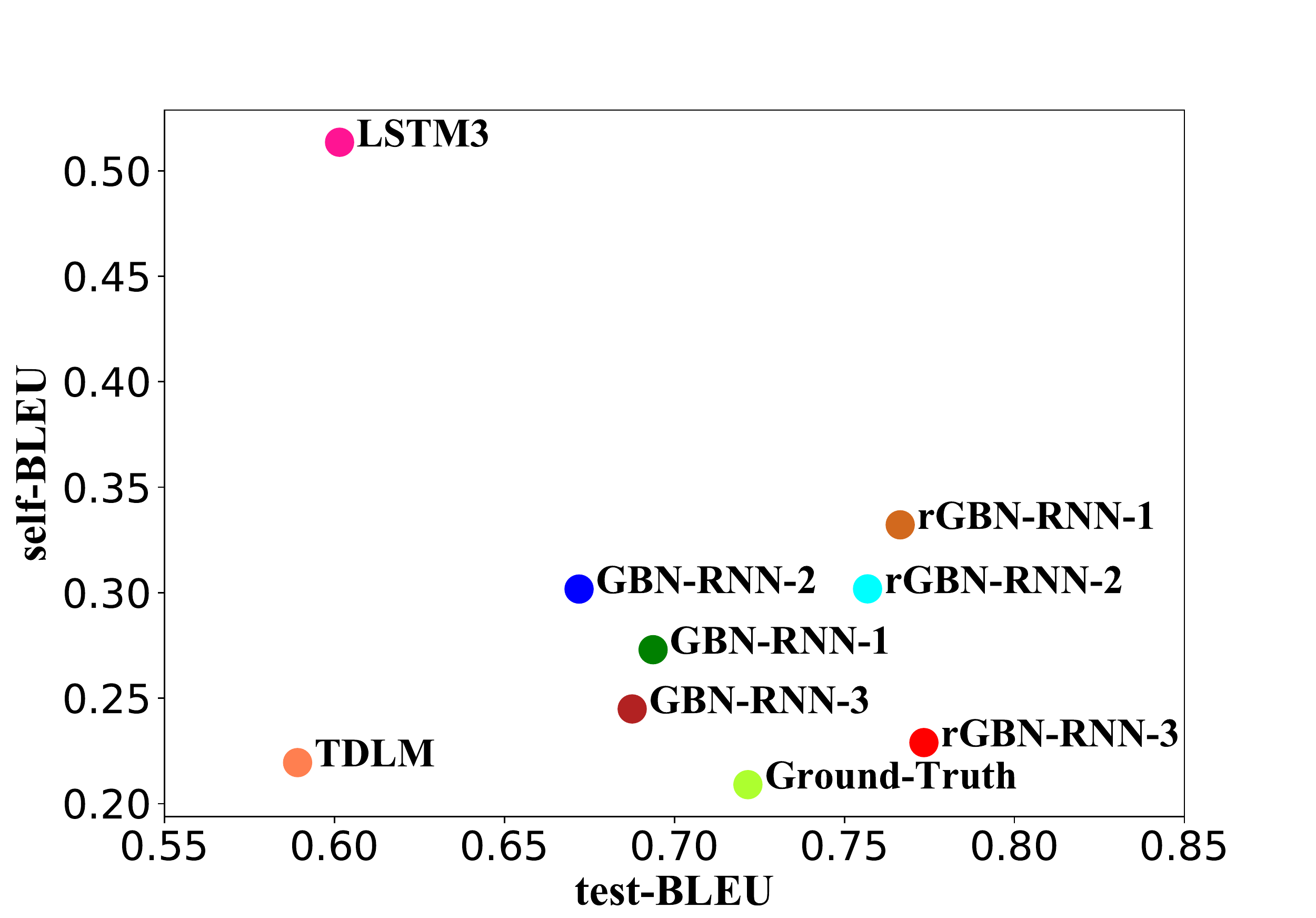}
\includegraphics[height=4.5cm,width=6.6cm]{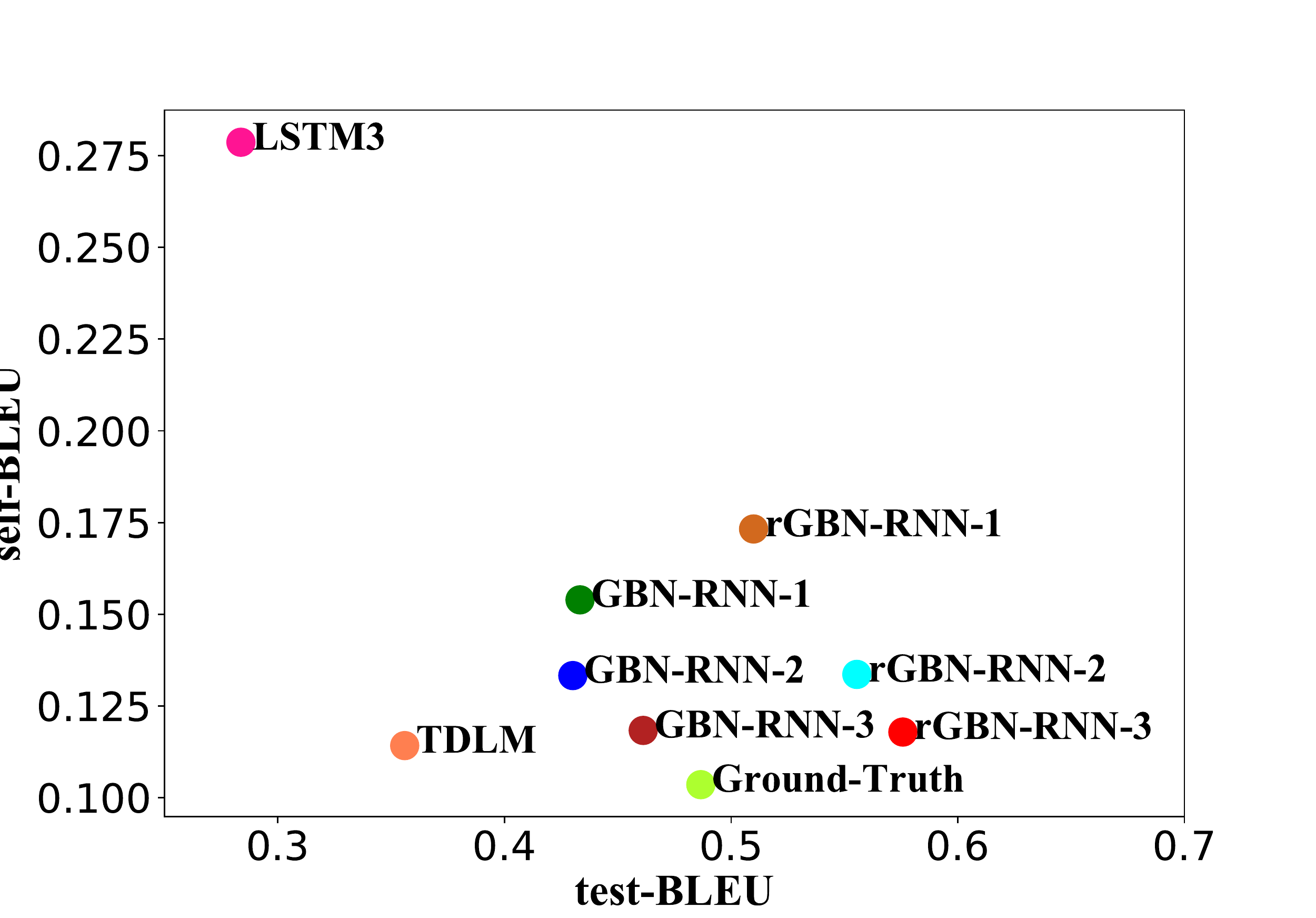}\vspace{-3mm}
\caption{ BLEU scores of different methods for IMDB. x-axis denotes test-BLEU, and y-axis self-BLEU. Left panel is BLEU-3 and right is BLEU-4, and a better BLEU score would fall within the lower right corner, where black point represents mean value and circles with different colors denote the elliptical surface of probability of BLEU in a two-dimensional space.}
\label{fig:imdb_BLEU} %
\end{center}
\end{figure*}

\section{BLEU scores for APNEWS}\label{sec:BLEU_APNEWS}\vspace{-2mm}
\begin{figure*}[ht!]
\begin{center}
\includegraphics[height=4.5cm,width=6.6cm]{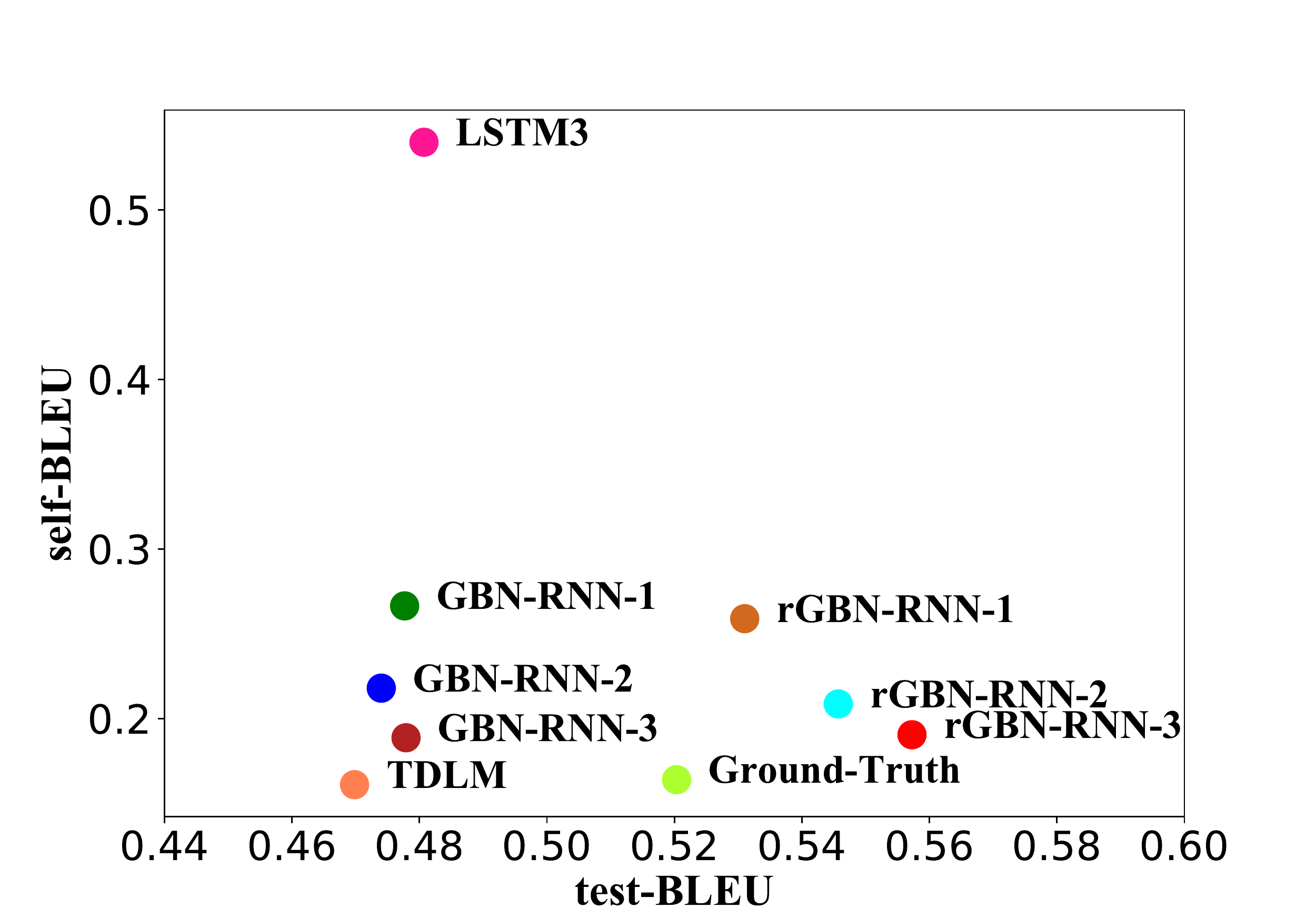}
\includegraphics[height=4.5cm,width=6.6cm]{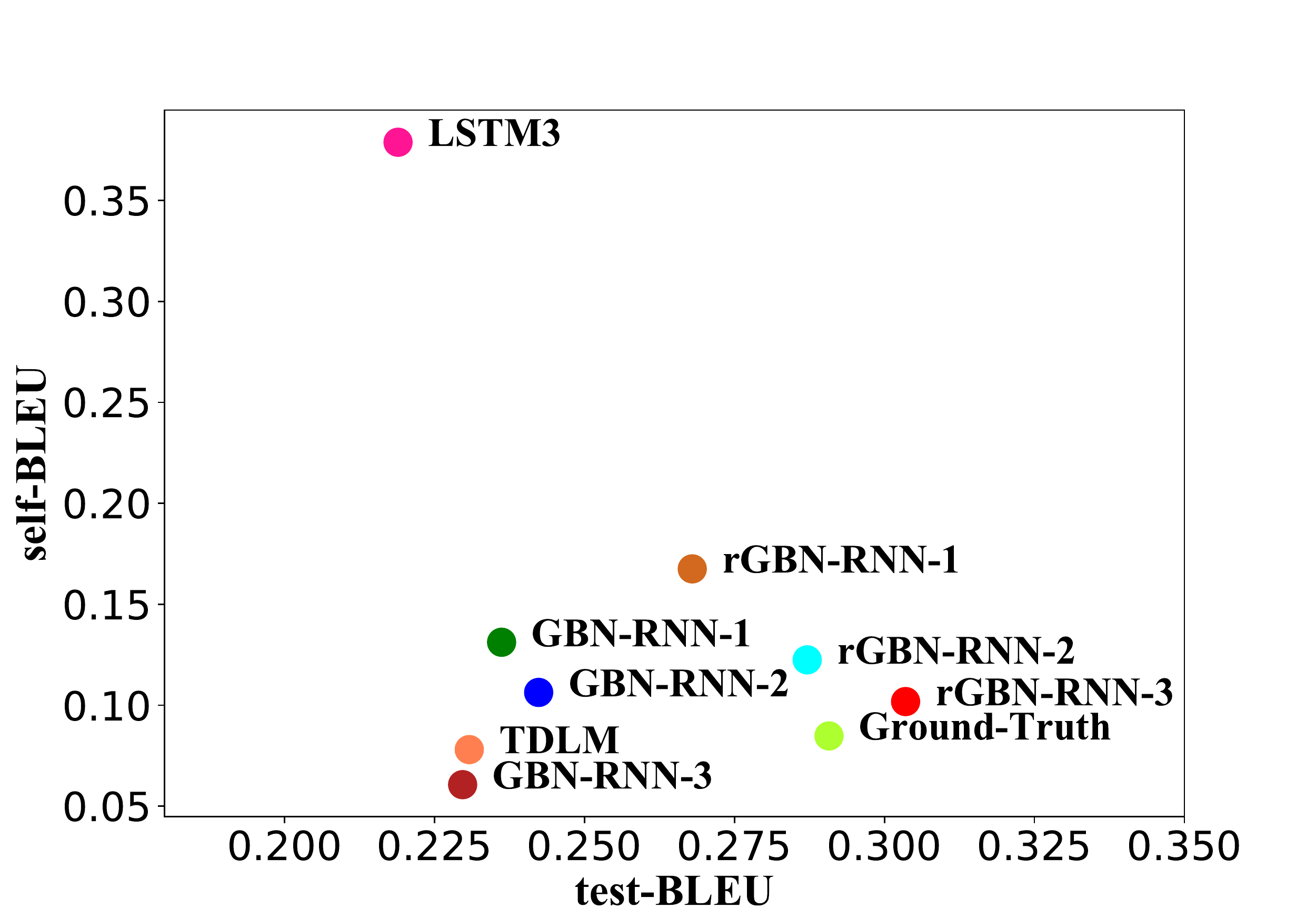}\vspace{-3mm}
\caption{ BLEU scores of different methods for APNEWS. x-axis denotes test-BLEU, and y-axis self-BLEU. Left panel is BLEU-3 and right is BLEU-4, and a better BLEU score would fall within the lower right corner, where black point represents mean value and circles with different colors denote the elliptical surface of probability of BLEU in a two-dimensional space.
}
\label{fig:apnews_BLEU} %
\end{center}
\end{figure*}

\clearpage
\section{Additional experimental results on IMDB and BNC}\label{sec:topics_and_sentences}

\begin{figure*}[!ht]
\begin{center}
\includegraphics[height=9.5cm,
]{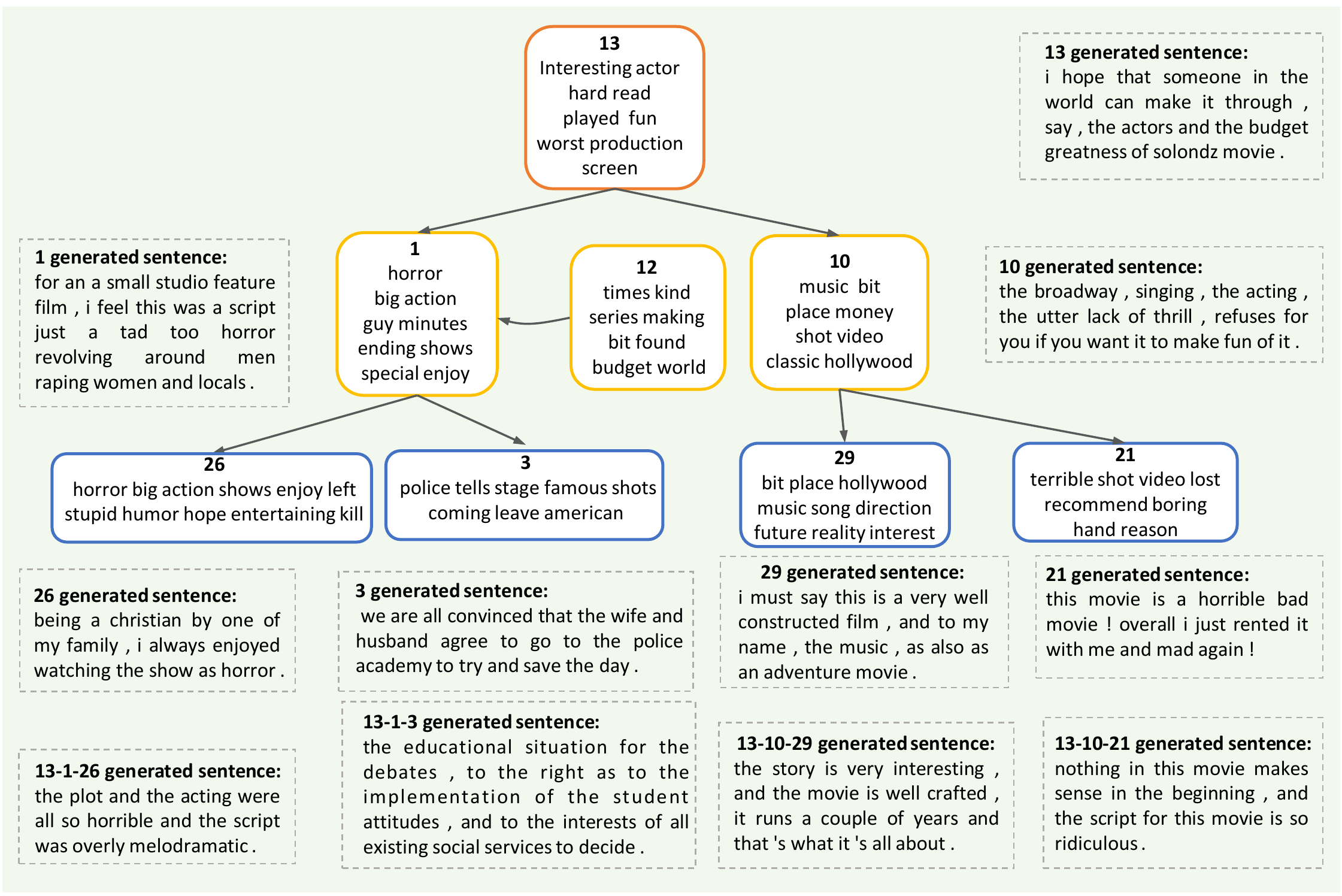}%
\caption{
Analogous plot to Fig. \ref{fig:sentence_topic} for the
IMDB corpus.
}
\label{fig:imdb_topics1} %
\end{center}
\end{figure*}

\begin{figure*}[!ht]
\begin{center}
\includegraphics[height=9.5cm,
]{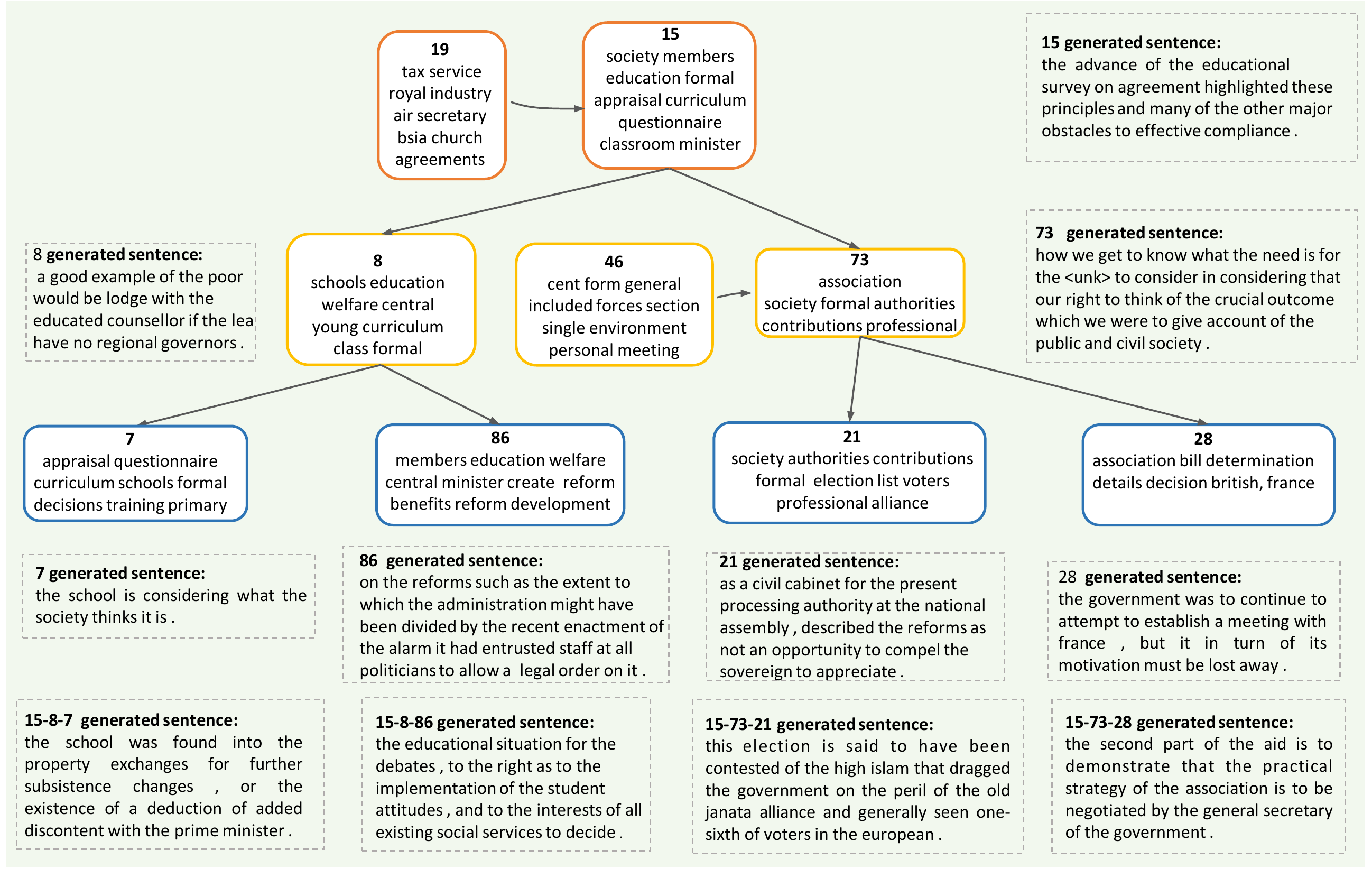}%
\caption{{
Analogous plot to Fig. \ref{fig:sentence_topic} for the
BNC corpus.
}
}
\label{fig:bnc_topics1} %
\end{center}
\end{figure*}

\clearpage
\section{{Additional example topic hierarchies and generated sentences.}}\label{sec:hierarchical_topics}

\begin{figure*}[!ht]
 \begin{center}
 \subfigure[]{
 \includegraphics[height=3.8cm,width=17.3cm]{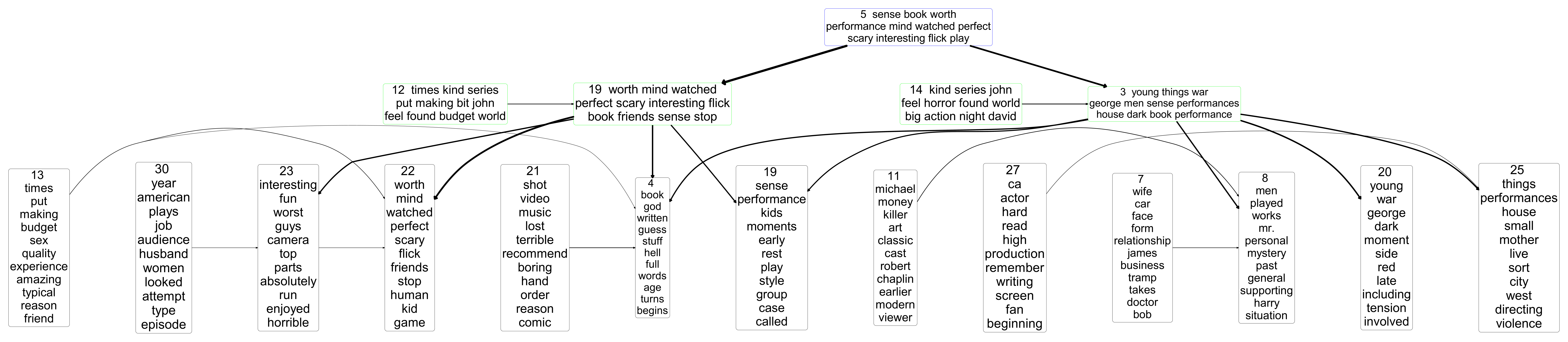}
 }
 \subfigure[]{
 \includegraphics[height=3.5cm,width=17.3cm]{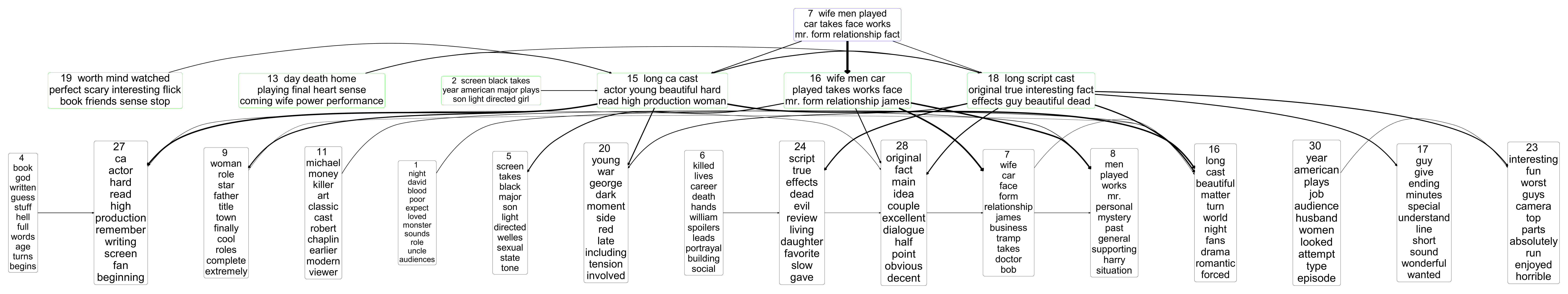}}
 \subfigure[]{
 \includegraphics[height=3.3cm,width=17cm]{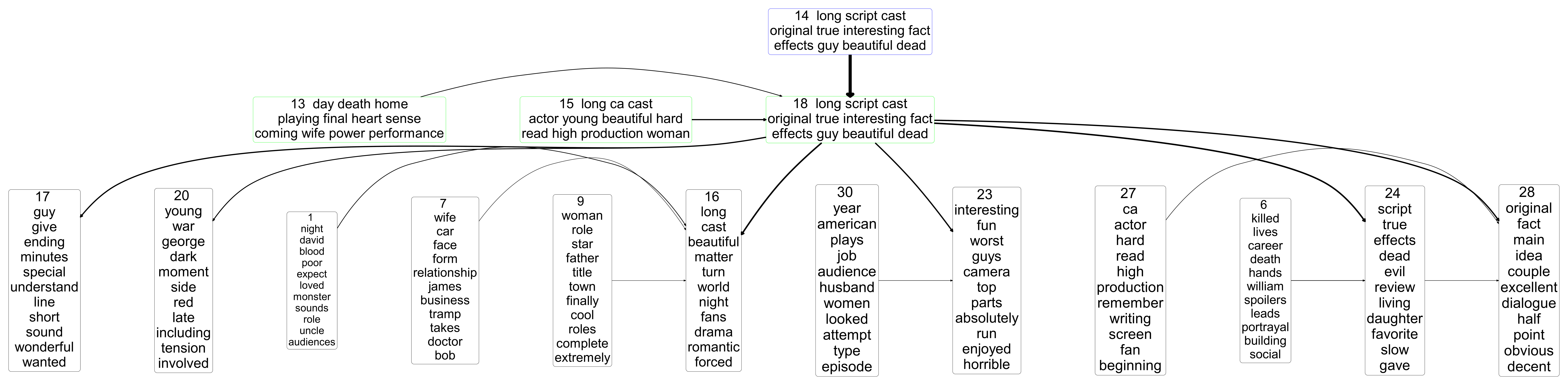}
 }\vspace{-4mm}
 \caption{{Example topics and their hierarchical and temporal connections inferred by a three-hidden-layer rGBN-RNN from the
IMDB corpus. Top words of each topic at layers 3, 2, and 1 are shown in blue, green, and black boxes respectively. Shown in (a)-(c) are the 5th, 7th and 14th nodes of the top layer, respectively.}}
\end{center} \vspace{-7mm}
\end{figure*}

\textit{\paragraph{Generated sentences conditioned on topic 5 at layer 3:} \small{(a) i love this movie , i strongly recommend it , just watch it with friends and laugh . (b) i was seriously shocked with hogan 's performance. (c) the performances are terrible , the storyline is non-existing , the directing ( if you can call it that ) is horrible , and the acting is horrible .}}
\textit{\paragraph{Generated sentences conditioned on topic 3 at layer 2:} \small{(a) her performance is a bit afro looking and they have a very neat southern accent and the movie was well cast as in previous movies , was so much more beautiful especially when woody allen made this film . (b) without their personal history or the fact that he would like to kill his american adoptive parents in their own homes , it all wo n't make sense . (c) this is one of those romantic comedies where we have some good action and good acting . }}
\textit{\paragraph{Generated sentences conditioned on topic 20 at layer 1:} \small{ (a) the new story was very touching , real , a new perspective of what it is like to be a back to war . (b) the movie is ok in my eyes , i know you will find it too scary ... but i kind of got a good movie from a somewhat dark sense . (c) at the same time i believe that the film was made just a couple of years earlier , the members of the u.s. government were almost always bad .}}
\textit{\paragraph{Generated sentences conditioned on a combination of topics 5, 19 and 22 at layer 3, 2, 1:} \small{(a) the show is well worth watching , i feel it is the best movie i have ever seen . (b) the movie is amazing , i thought the acting was great , and the movie was well worth the rental . (c) it would have been much better if it was a science fiction movie with a little more humor and some more action .}}

\begin{figure*}[!ht]
 \begin{center}
 \subfigure[]{
 \includegraphics[height=3.4cm,width=10cm]{figure/apnews_start3_node10_end1updown.pdf}
 }
 \subfigure[]{
 \includegraphics[height=3.6cm,width=17.3cm]{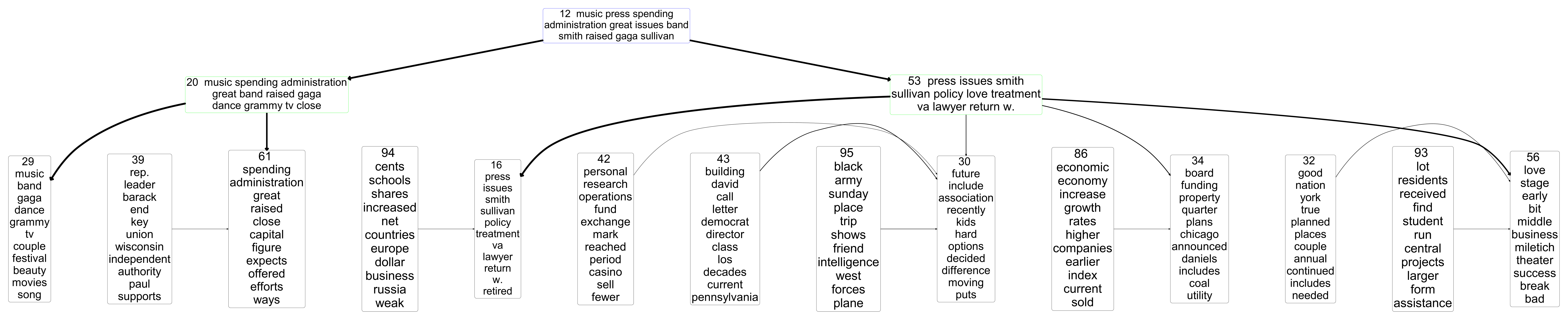}}
 \subfigure[]{
 \includegraphics[height=3.4cm,width=7cm]{figure/apnews_start3_node21_end1updown.pdf}
 }\vspace{-4mm}
 \caption{ {Example topics and their hierarchical and temporal connections inferred by a three-hidden-layer rGBN-RNN from the
APNEWS corpus. Top words of each topic at layers 3, 2, and 1 are shown in blue, green, and black boxes respectively. Shown in (a)-(c) are the 10th, 12th and 21th nodes of the top layer, respectively.}}
\end{center} \vspace{-5mm}
\end{figure*}
\textit{\paragraph{Generated sentences conditioned on topic 12 at layer 3:} \small{(a) they 're planning to attend a concert hall held by the rev. jesse. (b) approved by the standard free press , it will generate water for their own offices .
 (c) the christie administration will not give him an opinion if the of the state has issued its name . }}
\textit{\paragraph{Generated sentences conditioned on topic 53 at layer 2:} \small{(a) . the national park service said the maine department of law will hold a agreement on the <unk> law for the first time . (b) the detroit free press reports the city asked former winston-salem public schools commission chairman <unk> <unk> to take the seat . (c) earlier this month , the state police issued several orders to the fbi and send a <unk> team to the sheriff 's office . }}
\textit{\paragraph{Generated sentences conditioned on topic 29 at layer 1:} \small{(a) but it was and a few months later , the music had performed at the <unk> theatre in its <unk> . (b) the festival draws hundreds of thousands of viewers , a tourist year by a member of the oxford state team . (c) the university made the first " the most exciting , very beautiful " album followed by the 1996 " the sky " includes a <unk> version . }}
\textit{\paragraph{Generated sentences conditioned on a combination of topics 21, 46 and 44 at layer 3, 2, 1:} \small{(a) police say the suspect was taken to a hospital for treatment . (b) the man was arrested after police say he was driving in a car in north mississippi , which was the first <unk> to be used in the shootout . (c) police said wednesday that the victims ' deaths are not believed to be gang affiliation . }}

\begin{figure*}[!ht]
 \begin{center}
 \subfigure[]{
 \includegraphics[height=3.7cm,width=13cm]{figure/bnc_start3_node1_end1updown.pdf}
 }
 \subfigure[]{
 \includegraphics[height=3.5cm,width=17.3cm]{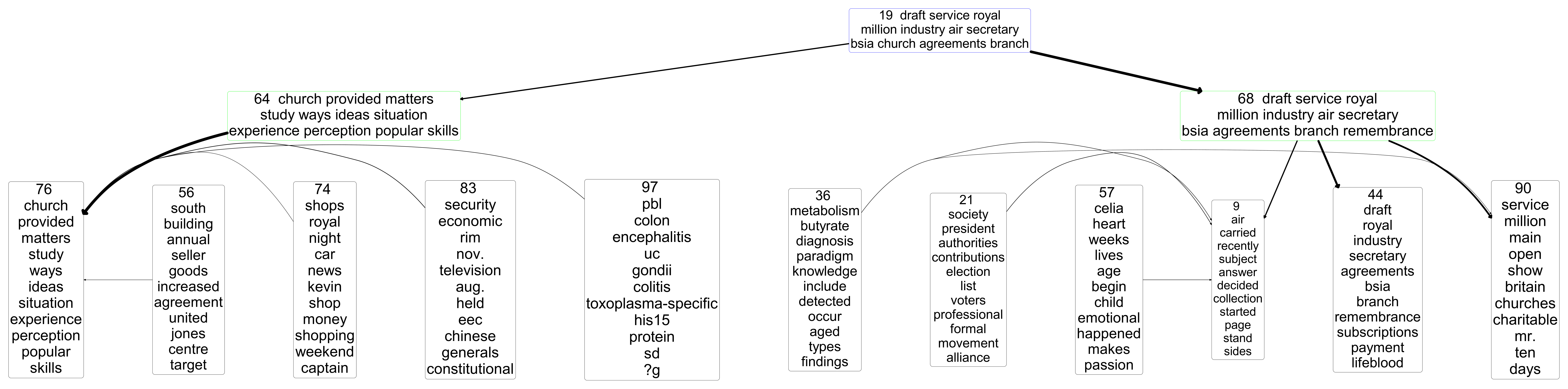}}
 \subfigure[]{
 \includegraphics[height=3.6cm,width=17.3cm]{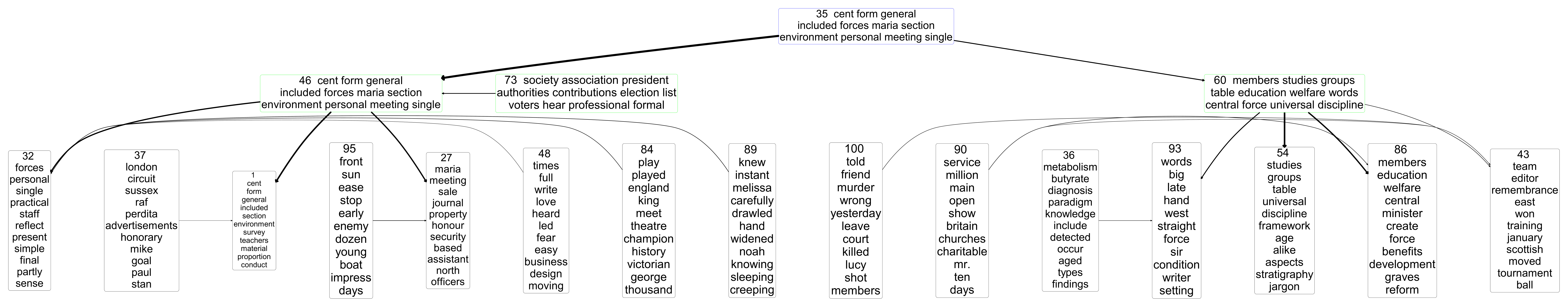}
 }\vspace{-4mm}
 \caption{ {Example topics and their hierarchical and temporal connections inferred by a three-hidden-layer rGBN-RNN from the
BNC corpus. Top words of each topic at layers 3, 2, and 1 are shown in blue, green, and black boxes respectively. Shown in (a)-(c) are the 15th, 19th and 35th nodes of the top layer, respectively.}}
\end{center} \vspace{-5mm}
\end{figure*}
\textit{\paragraph{Generated sentences conditioned on topic 1 at layer 3:} \small{(a) the fourth should be in the obligation of the enforcement officer for proof where a supplementary liability of pension funds has not been advocated. (b) approved by the standard free press , it will generate water for their own offices . (c) the court , has recently agreed to participate in the investigation to allow the justices to succeed to be responsible for the full remit of the submissions . }}
\textit{\paragraph{Generated sentences conditioned on topic 73 at layer 2:} \small{(a) another , which was the period of message from a federal of the new presidential opposition to the new constitution , was to change his autonomy rather than through the different strategies .
 (b) the president is to bring out the best of all and the most widely understood state of affairs in the country . (c) the icrc has announced that the <unk> could not be accepted on the factors outlined in the societies ' choice . }}
\textit{\paragraph{Generated sentences conditioned on topic 76 at layer 1:} \small{(a) the church of st clement danes -- the great continent , the southern of the realm , a treaty give such assistance to brother. (b) the legality of the political instability that followed by the profession has been criticised for the necessity for the study of individuals and friends . (c) the character of the monarchy is dominated by a panoramic style which includes and attempts to limit the genre to his/her ideas . ' }}
\textit{\paragraph{ Generated sentences conditioned on a combination of topics {19, 68 and 44} at layer 3, 2, 1:} \small{(a) again the royal air force in the middle of the war was now part of the struggle by the indian resistance . (b) the mailing list for another example of the british aerospace industry shows a, exclusive catalogue to enable object to be changed to steam . (c) the great britain will continue to sumbit to the thinking and nature of the reciprocal international economic agreement.}}

\section{{Additional examples of generated sentences / paragraph conditioning on a paragraph.}}\label{sec:paragraphs}

\begin{figure}[!ht]
\begin{center}
\includegraphics[%
width=.95\hsize]{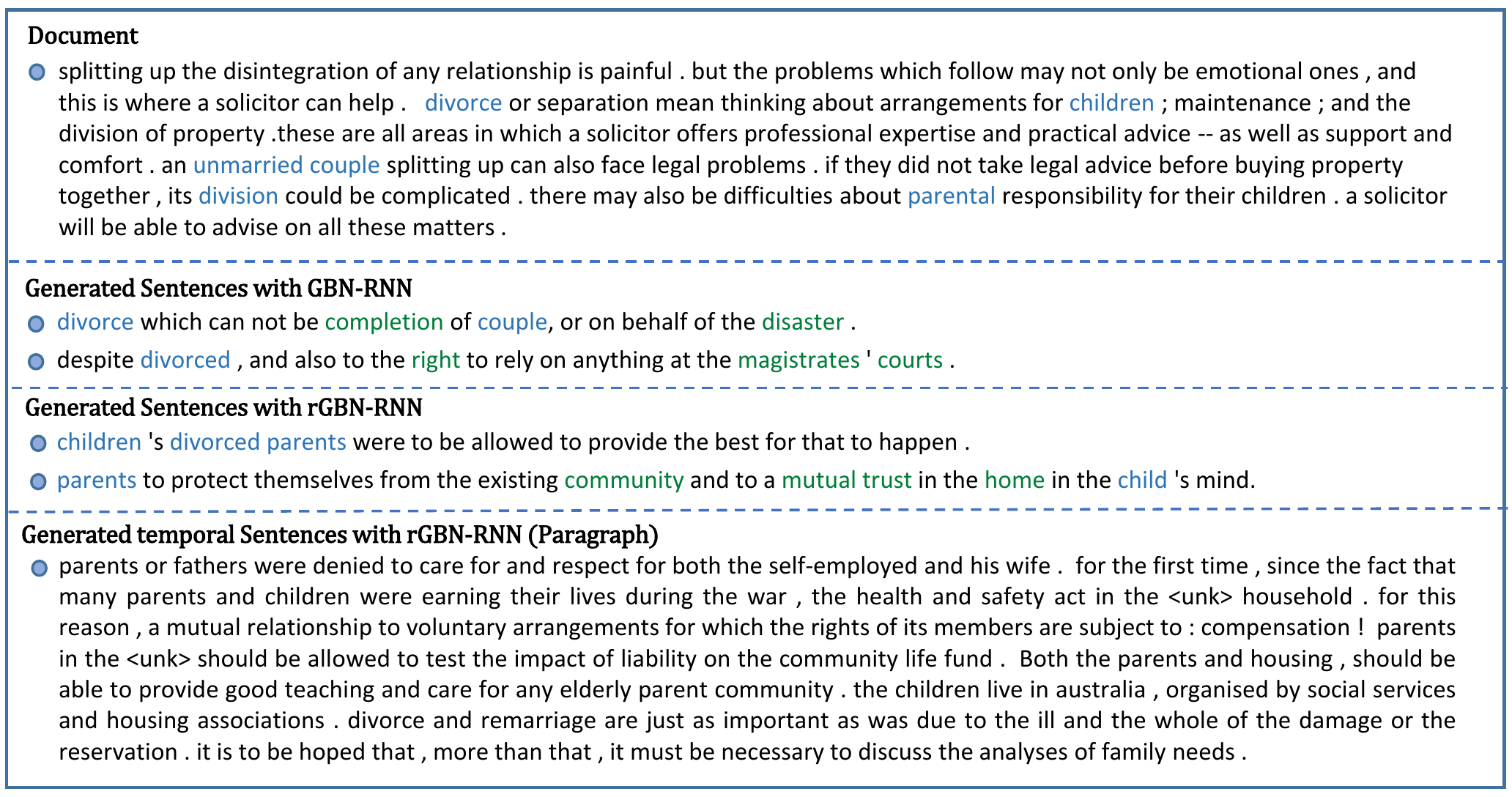}\vspace{-3mm}
\caption{ An example of generated sentences and paragraph conditioned on a document from BNC (green denotes novel words, blue the key words in document and generated sentences.)
}
\label{fig:sentence_document_bnc} \vspace{-3mm} %
\end{center}
\end{figure}

\begin{figure}[!ht]
\begin{center}
\includegraphics[%
width=.95\hsize]{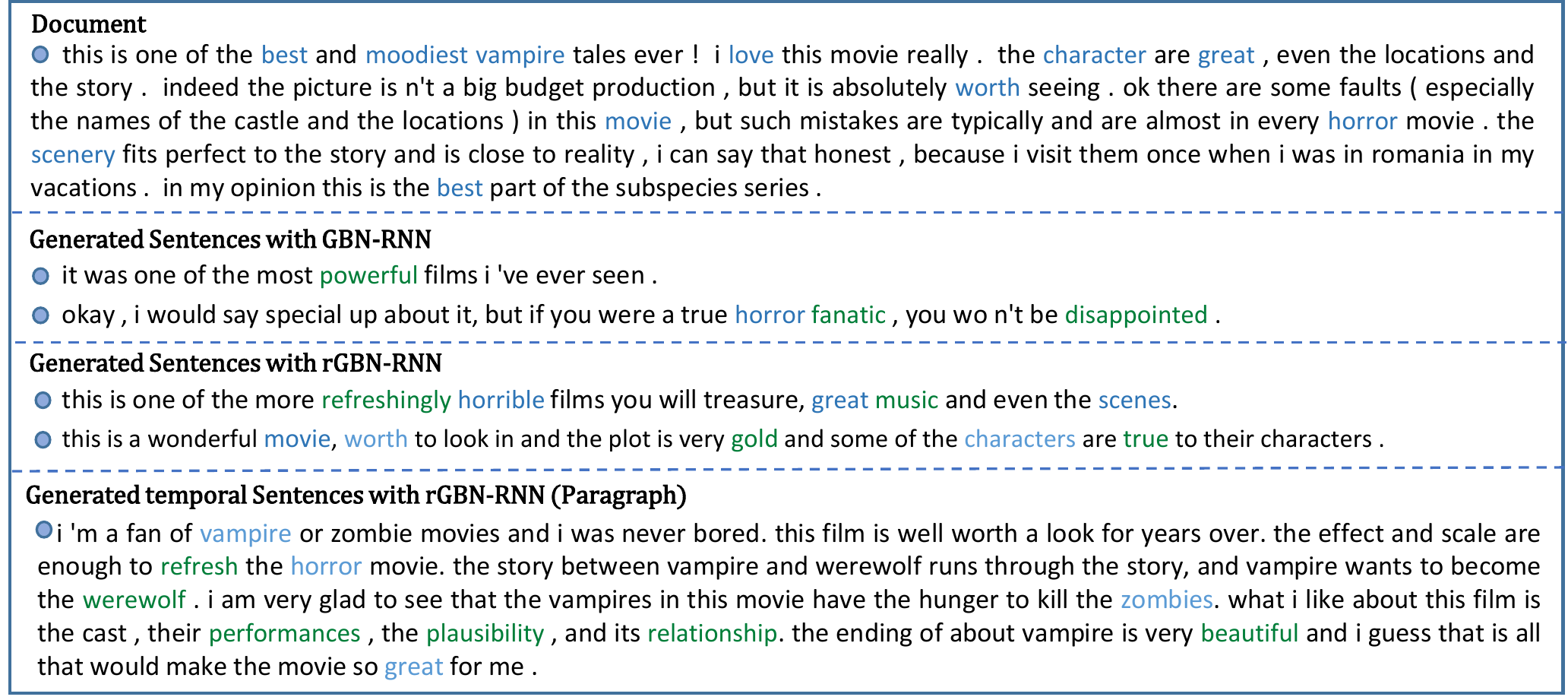}\vspace{-3mm}
\caption{\small An example of generated sentences and paragraph conditioned on a document from IMDB (green denotes novel words, blue the key words in document and generated sentences.)
}
\label{fig:sentence_document_imdb} \vspace{-3mm} %
\end{center}
\end{figure}

\end{document}